\documentclass[journal]{IEEEtran}

\usepackage{rotating}
\usepackage[hidelinks]{hyperref} 
\usepackage{multirow}
\usepackage{diagbox}
\usepackage{graphicx}
\usepackage{subcaption}
\usepackage{siunitx}
\usepackage[linesnumbered,ruled,vlined,noend]{algorithm2e}
\usepackage{soul}
\usepackage{balance}
\usepackage{booktabs}
\usepackage{amssymb}
\usepackage{amsmath}
\usepackage{tikz}

\newcommand{\myNum}[1]{(\emph{#1})}
\newcommand{\circlednum}[1]{\raisebox{.5pt}{\textcircled{\raisebox{-.9pt}{#1}}}}

\newcommand{\eat}[1]{}
\newcommand{\stab}{\rule{0pt}{8pt}\\[-2.5ex]}
\newcommand{\eg}{{\it e.g.}, }
\newcommand{\ie}{{\it i.e.}, }

\newcommand{\GraphCom}{{\sc GraphComp}}
\newcommand{\GNN}{{\sc T-AutoG}}
\newcommand{\MAbsD}{{\sc MAbsD}}

\newcommand{\guozhong}[1]{\textcolor{red}{#1}}
\let\guozhong\relax

\def\RS{\textsc{RedSea}}
\def\ERA{\textsc{Era}}
\def\Miranda{\textsc{Miranda}}
\def\SCALE{\textsc{Scale}}
\def\WGAN{\textsc{rsSyn}}
\def\CESM{\textsc{Cesm}}
\def\calO{\mathcal{O}}

\newtheorem{definition}{Definition}

\SetKwComment{Comment}{{// }}{}

\SetCommentSty{mycommfont}

\begin{document}

\title{Extreme Error-bounded Compression of Scientific Data via Temporal Graph Autoencoders}

\author{\thanks{Manuscript received XX XX, XXXX}
  Guozhong~Li,
  Muhannad Alhumaidi,
  Spiros Skiadopoulos,
  Ibrahim Hoteit,
  and~Panos Kalnis
\IEEEcompsocitemizethanks{\IEEEcompsocthanksitem Guozhong Li, Muhannad Alhumaidi, and Panos Kalnis are with the Computer, Electrical and Mathematical Sciences and Engineering Division, KAUST, Thuwal, Saudi Arabia. \protect
E-mail: \{guozhong.li, muhannad.humaidi, panos.kalnis\}@kaust.edu.sa
\IEEEcompsocthanksitem Spiros Skiadopoulos is with the Department of Informatics and Telecommunications, University of Peloponnese, Tripoli, Greece. 
E-mail: spiros@uop.gr
\IEEEcompsocthanksitem Ibrahim Hoteit is with the Physical Science and Engineering Division,
 KAUST, Thuwal, Saudi Arabia. 
E-mail: ibrahim.hoteit@kaust.edu.sa}
\thanks{For computer time, this research used IBEX and Shaheen III, managed by the Supercomputing Core Laboratory at King Abdullah University of Science and Technology (KAUST), Saudi Arabia.}
}

\maketitle 

\setlength{\abovecaptionskip}{0.05\baselineskip}
\setlength{\belowcaptionskip}{0.05\baselineskip}

\begin{abstract}
   The generation of voluminous scientific data poses significant challenges for efficient storage, transfer, and analysis.
   Recently, error-bounded lossy compression methods emerged  due to their ability to achieve high compression ratios while controlling data distortion.
   However, they often overlook the inherent spatial and temporal correlations within scientific data, thus missing opportunities for higher compression. 
   In this paper we propose \GraphCom, a novel graph-based method for error-bounded lossy compression of scientific data.
   We perform irregular segmentation of the original grid data and generate a graph representation that preserves the spatial and temporal correlations. Inspired by Graph Neural Networks (GNNs), we then propose a temporal graph autoencoder to learn latent representations that significantly reduce the size of the graph, effectively compressing the original data. Decompression reverses the process and utilizes the learnt graph model together with the latent representation to reconstruct an approximation of the original data. The decompressed data are guaranteed to satisfy a user-defined point-wise error bound.   
   We compare our method against the state-of-the-art error-bounded lossy methods (i.e., HPEZ, SZ3.1, SPERR, and ZFP) on large-scale real and synthetic data.
   \GraphCom~consistently achieves the highest compression ratio across most datasets, outperforming the second-best method by margins ranging from 22\% to 50\%.
\end{abstract}



\section{Introduction}\label{sec:introduction}
\IEEEPARstart{M}{odern} sensing and simulation technologies generate large volumes of scientific data, in the order of hundreds of terabytes to petabytes, across different domains such as weather and climate~\cite{hoteit-RSRA2018,kay2015community,hoteit-RSRA2022}, geopotential simulations~\cite{hersbach2020era5}, and seismic wave analysis~\cite{kayum2020geodrive}.
The large data volume poses significant challenges on storage, network and computational infrastructures~\cite{gray2005scientific}.  
To address these challenges, advanced data reduction techniques for the compression of large scientific data become essential for efficient storage, transfer, and analysis.

Scientific data libraries (\eg NetCDF~\cite{rew1990netcdf} and HDF5\footnote{\url{http://www.hdfgroup.org/HDF5}}) implement traditional \emph{lossless} compression techniques, such as Zlib\footnote{\url{http://www.zlib.net/}} and Zstandard\footnote{\url{https://github.com/facebook/zstd/}}, that guarantee data fidelity, but only achieve low compression ratios, rarely exceeding 2$\times$~\cite{zhao2020sdrbench}.
At the other end of the spectrum, there exist \emph{error-unbounded lossy} compression methods~\cite{klema1980singular,jolliffe2016principal,huang2023compressing} that promise impressive compression ratios but may introduce large errors, thus compromising the accuracy and utility of the decompressed data.

\begin{figure}[t]
    \centering
    \begin{subfigure}[b]{0.45\linewidth}
        \centering
        \includegraphics[width=\linewidth]{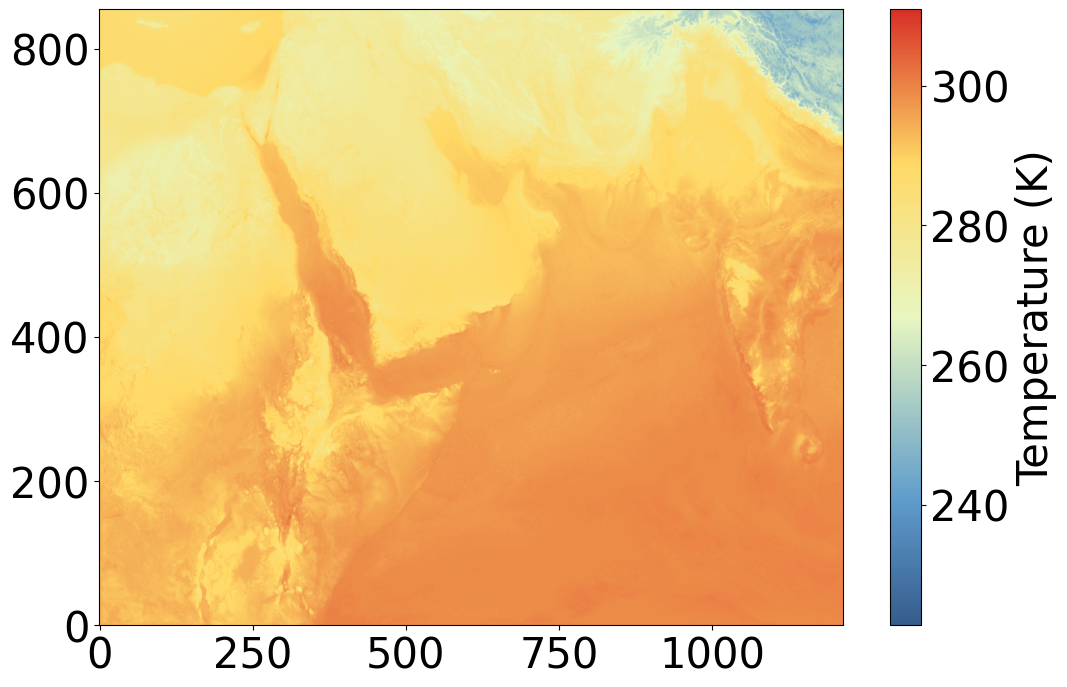}
        \caption{A single timestamp}
        \label{fig:one-specific-timestamp-from-redsea500}
    \end{subfigure}
    \begin{subfigure}[b]{0.5\linewidth}
        \centering
        \includegraphics[width=.93\linewidth]{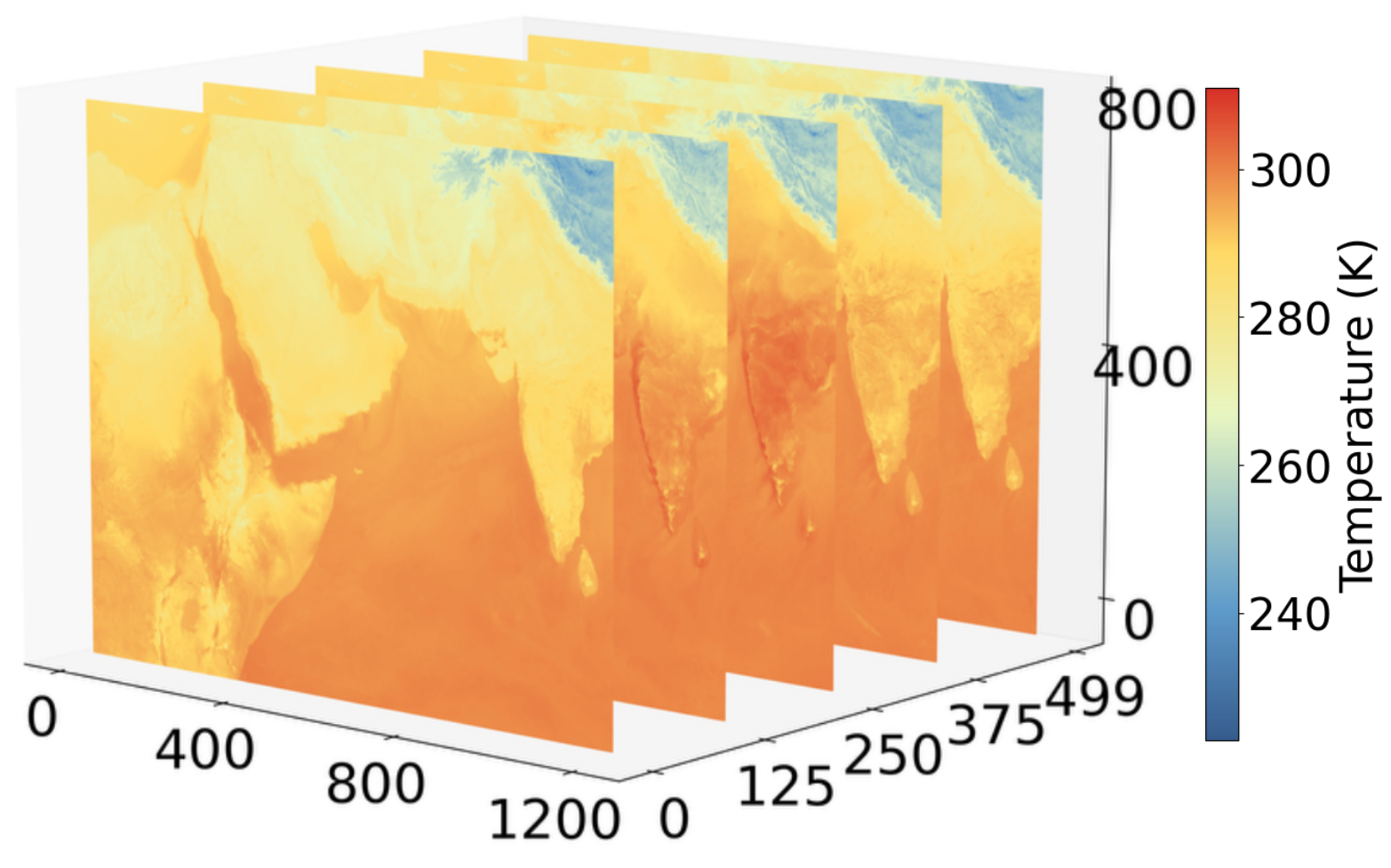} 
        \caption{Time series of 500 timestamps}
        \label{fig:original-redsea500-data}        
    \end{subfigure}
    \caption{The {\RS} dataset: a visualization of Red Sea reanalysis temperature data~\cite{hoteit-RSRA2018} covering Eastern Africa, the Arabian Peninsula, the Indian Ocean and neighboring regions.}
    \label{fig:redsea-temp-example}
\end{figure}

The consensus of the scientific community is to utilize \emph{error-bounded lossy} compression to achieve better compression ratios than lossless methods, while ensuring some error metric appropriate for the particular application domain (\eg relative point-wise error) to be within a user-defined bound $\epsilon$.   
Several error-bounded lossy compressors have demonstrated good performance in handling scientific data\footnote{There are many types of scientific data, including numerical, categorical, text, and image data; this paper specifically focuses on \emph{numerical} data.}, including HPEZ~\cite{liu2023high}, SZ3.1~\cite{zhao2021optimizing}, SPERR~\cite{li2023lossy}, and ZFP~\cite{lindstrom2014fixed}.
Meanwhile, recent developments in deep neural networks resulted in NN-based approaches, such as autoencoder-based SZ~\cite{liu2021exploring}, coordinate network-based compressors~\cite{han2022coordnet}, and super-resolution network-based SZ~\cite{liu2023srn}.

\begin{figure*}[tbp]
  \centering
  \includegraphics[width=\linewidth]{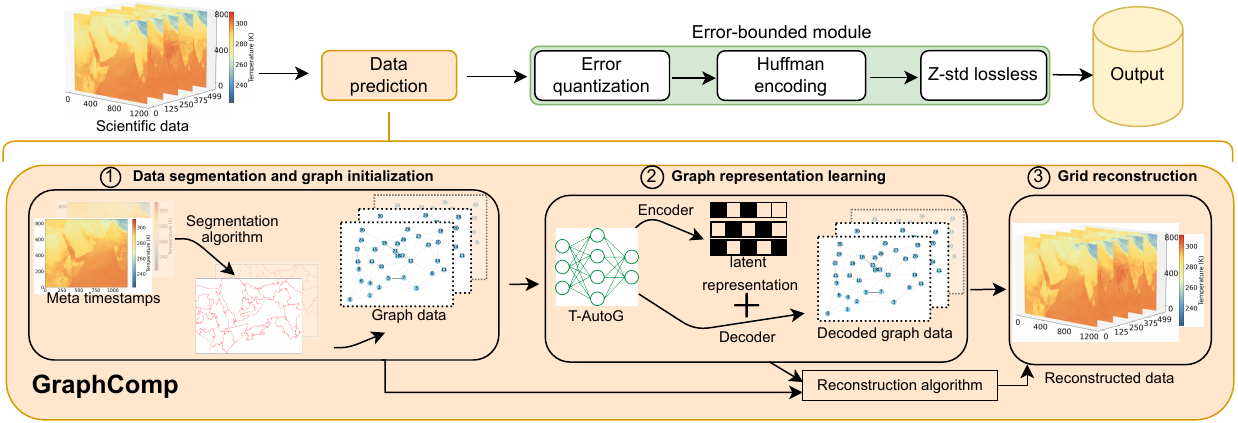}
  \caption{Overview of \GraphCom, which includes three primary steps: \circlednum{1} graph initialization, \circlednum{2} graph representation learning, and \circlednum{3} grid reconstruction. To guarantee a point-wise error bound, we further integrate an error-bounded module.}\label{fig:workflow}
\end{figure*}

\stab
\noindent{\bf Challenges.}
Despite advancements in error-bounded lossy compression, significant challenges remain.
Existing methods often fail to adequately consider the spatial correlations within scientific data.
For instance, regional temperature data, as depicted in the example of Figure~\ref{fig:one-specific-timestamp-from-redsea500}, exhibit complex spatial patterns where nearby elements frequently express similar attributes.
Compression methods, such as HPEZ~\cite{liu2023high} and SZ3.1~\cite{zhao2021optimizing}, may fail to accurately capture such irregular relationships, thus compromising the potential for optimal spatial correlation exploitation for compression. 

Moreover, the temporal dynamics inherent to scientific datasets are often overlooked.
Temporal correlations, as illustrated in consecutive timestamps of the example in Figure~\ref{fig:original-redsea500-data}, require explicit modeling to ensure the dependencies are preserved during compression. 
Existing methods such as SPERR~\cite{li2023lossy} and ZFP~\cite{lindstrom2014fixed}, may introduce distortions to such temporal patterns, effectively lowering the compression ratio.

\stab
\noindent{\bf Our solution.}
We propose \GraphCom, a novel graph-based error-bounded lossy compression method for scientific data, that leverages temporal graph autoencoders.
Unlike regular grid structures, which may fail to capture complex nonlinear correlations, graphs excel in  representing irregular spatial relationships.
Temporal graphs enhance this capability by capturing temporal dynamics, enabling \GraphCom~to effectively compress complex scientific data that exhibit both spatial and temporal correlations.

An overview of \GraphCom~is presented in Figure~\ref{fig:workflow}.
Intuitively, scientific data consist of regions that contain roughly homogeneous values but may be of irregular shape. {\GraphCom} employs the Felzenszwalb~\cite{felzenszwalb2004efficient} algorithm, a well-known method for image segmentation, to partition each timestamp of the original data. 
The partitioning is then transform into a graph, where each region corresponds to a node and there is an edge between nodes if the corresponding regions are adjacent. The graph is augmented with aggregate information about the regions. 
The resulting series of graphs for all timestamps, depicted in Figure~\ref{fig:workflow}\circlednum{1}, reliably preserve both spatial and temporal correlations across the data.

The graph series forms the input to our proposed temporal graph autoencoder (named {\GNN}) for learning latent representations.
{\GNN} utilizes graph convolutional layers to learn the spatial information from graph data, transforming them into graph embeddings.
Subsequently, the embeddings are processed by convolutional layers to capture temporal information, producing latent representations that capture complex node interactions and their temporal dynamics, as shown in  Figure~\ref{fig:workflow}\circlednum{2}. 

The latent representations together with the learnt graph models are used in a reversed process to reconstruct the grid data, shown in Figure~\ref{fig:workflow}\circlednum{3}. Obviously, the reconstructed data are an approximation of the original data. To guarantee that the point-wise relative error is within a user-defined bound $\epsilon$, we compute and store the residual errors in a compact form. This information is used during decompression to correct any erroneously reconstructed data. We utilize a combination of error-controlled quantization, Huffman encoding, and lossless compression of the quantized data, following existing approaches \cite{liu2023high,zhao2021optimizing}. Consequently, {\GraphCom} transforms the original data to a compressed version consisting of the following components: \myNum{i} a compressed form of the segmentation; \myNum{ii} the weights of {\GNN}; \myNum{iii} a compressed form of the latent representations; and \myNum{iv} a compressed form of the residual errors. The total size (in bytes) of the compressed version is much smaller than the size of the original data.

We conduct comprehensive experiments on both real (including the private \RS reanalysis dataset \cite{hoteit-RSRA2018, hoteit-RSRA2022}, and public \ERA5 data~\cite{hersbach2020era5}), as well as synthetic data.
The results show that the compression ratios of our {\GraphCom} method are, in most cases, higher than the compared state-of-the-art methods (\ie  HPEZ, SZ3.1, SPERR, and ZFP), under all tested error bounds.
Compared to the second-best method, {\GraphCom} improves compression ratios from 22\% up to 50\%.

Note that the spatial resolution of the RSRA dataset is higher \cite{hoteit-RSRA2018, hoteit-RSRA2022} than that of the publicly available ones (\eg ERA5); the finer resolution puts additional stress on the compressors, revealing their inefficiencies. Since RSRA is private, for reproducibility and to serve the scientific community, we develop a synthetic data generator based on generative adversarial networks and trained on the RSRA data. We show that the behaviour of the compressors on the synthetic data is similar to that on the real data. Both the data generator and the {\GraphCom} code are available as open-source.

\stab
Our contributions are summarized as follows:
\begin{itemize}
    \item We propose \GraphCom, a graph-based error-bounded lossy compression method that results in high compression ratios of scientific data while preserving data integrity.
    \item We develop \GNN, a temporal graph autoencoder designed to capture both spatial and temporal information from the transformed graph data, facilitating latent representation learning and grid data reconstruction.
    \item We train a generative adversarial model to generate synthetic data that preserve the statistical properties of the proprietary RSRA data, providing a valuable resource for the community.
    \item We conduct extensive experiments on real and synthetic datasets, to demonstrate the superiority of {\GraphCom} against state-of-the-art compression techniques.
\end{itemize}

\eat{\noindent{\bf Organization.}
The remainder of this paper is organized as follows.
We introduce the background in Section~\ref{sec:preliminary}.
The details of our proposed method are given in Section~\ref{sec:method}.
Section~\ref{sec:experiment} reports the experimental results.
Section~\ref{sec:Related Work} reviews the related work.
Section~\ref{sec:conclusion} concludes the paper and presents the future work.}


\section{Background}\label{sec:preliminary}
This section presents the problem formulation; 
Table~\ref{tab:terminologies} summarizes some frequently used notations and their meanings.

\stab
\noindent \textbf{Temporal scientific dataset $X$.} 
A temporal scientific dataset that captures events across $T$ timestamps,  where each timestamp is a matrix with $M$ rows and $N$ columns, is represented as $X \in \mathbb{R}^{T\times M\times N}$.  
Figure~\ref{fig:redsea-temp-example} illustrates an example of the Redsea-500 dataset that includes 500 timestamps with temperature data from the Red Sea region.

\stab
\noindent \textbf{Compression ratio $\rho$.}
Let $|X|$ be the the size of the original data in bytes. 
Let $|Z|$ be the total size of the compressed data in bytes; this includes all the information that is needed for decompression, such as any metadata, segmentation regions, model weights, latent representations, etc. Compression ratio is defined as: 
\begin{equation}\label{eq:compression_ratio}
    \rho = \frac{|X|}{|Z|}
\end{equation}

\stab
\noindent \textbf{Problem formulation.} 
Let \( X \in \mathbb{R}^{T \times M \times N} \) be a temporal scientific dataset and $\epsilon$ be a user-specified relative error bound\footnote{The absolute error bound \( e \) can be derived from the relative error bound \( \epsilon \), as \( e = \epsilon \cdot \mathrm{vrange}(X) \).}. 
Let $C$ be a compression function that generates the compressed version $Z$ of $X$, \ie 
$Z=C(X)$. Let $D$ be a decompression function that generates the decompressed version $X'$  of $Z$, \ie $X' = D(Z)$. The process is lossy, therefore $X'$ is an approximation of $X$.

The goal of this paper is to define a pair of functions $\langle C, D\rangle$, such that the compression ratio $\rho$ of $X$ is maximized, while the relative difference between each original data point $X_i$ and its decompressed counterpart $X'_i=D(C(X))_i$ does not exceed the relative error bound \( \epsilon \). Formally: 
\begin{equation}\label{eq:formulation}
    \underset{\langle C, D\rangle}{\arg\max} \frac{|X|}{|C(X)|}\ \ 
    \text{such that }  \frac{|X_i - D(C(X))_i|}{\mathrm{vrange}(X)} \le \epsilon, 
\end{equation}
where $\mathrm{vrange}$ returns the range of values in a dataset.


\begin{table}[t]
  \centering
  \caption{Summary of frequently used notation}
    \begin{tabular}{cp{3.3cm}||cp{3.1cm}}
    \hline
    $X$            & temporal scientific dataset $\in \mathbb{R}^{T\times M\times N}$ &
    $T$            & number of timestamps in $X$ \\
    $M$            & number of rows in $X$ &
    $N$            & number of columns in $X$ \\
    $\epsilon$     & user-specified relative error bound &
    $e$            & user-specified absolute error bound \\
    $C$            & compressor &
    $\rho$         & compression ratio \\ 
    $D$            & decompressor &
    $Z$            & compressed data \\
    $X'$           & decompressed data &
    $\theta$       & parameters in segmentation algorithm \\
    $G$            & graph $G = (V, E, F)$ &
    $F$            & node features $F \in \mathbb{R}^{|V|\times d}$ \\
    $V$            & set of nodes &
    $E$            & set of edges \\
    $A$            & adjacency matrix of graph $G$ &
    $\mathcal{G}$  & graph set $\mathcal{G} = \bigcup_{i=1}^{T}G_i $ \\ 
    \hline
    \end{tabular} 
  \label{tab:terminologies}%
\end{table}%

%

\section{Related Work}\label{sec:Related Work}

\stab\noindent
{\bf Scientific data compression.}
Recent advances in scientific simulation and sensing led to the generation of voluminous data across diverse fields such as weather and climate~\cite{kay2015community,hoteit-RSRA2022,hersbach2020era5}, seismic wave analysis~\cite{kayum2020geodrive}, and turbulence studies\footnote{\url{https://wci.llnl.gov/simulation/computer-codes/miranda}}. 
The proliferation of data poses significant challenges in terms of efficient storage and transfer. 
While lossless compression techniques are prevalent in database management systems~\cite{cudre2009demonstration,yu2020two, pelkonen2015gorilla}, their application to scientific data is often limited by modest compression ratios\footnote{For example, Gorilla~\cite{pelkonen2015gorilla}, an in-memory database with lossless compression, reports up to $10\times$ compression ratio for time series. However, on the Redsea data the compression ratio is less than $2\times$.}, typically less than $2\times$~\cite{zhao2020sdrbench}.
In response to the limitation, there has been a surge to explore lossy compression methods.
The primary challenge lies in achieving maximal compression while bounding the error, thereby ensuring the integrity of data decompression for subsequent analysis.

\stab\noindent
{\bf Error-bounded lossy compressors.}
We focus on error-bounded lossy compressors, where the quality of decompressed data remains within specified error tolerances.
ZFP~\cite{lindstrom2014fixed} offers a fixed-bitrate mode through the truncation of blocks of orthogonally transformed data.
SPERR~\cite{li2023lossy} is a lossy compressor for structured scientific data, incorporating the advanced wavelet compression algorithm SPECK~\cite{pearlman2004efficient,tang2006three}.
SZ-based methods, including SZ2+~\cite{tao2017significantly}, SZ3~\cite{liang2022sz3}, SZauto~\cite{zhao2020significantly}, and HPEZ~\cite{liu2023high}, are among the most popular error-bounded lossy compressors.
They typically involve four main steps: data prediction, error quantization, Huffman encoding, and lossless residual error compression. 
SZ-based methods have evolved in their approach to prediction, ranging from simple one-dimensional adaptive curve-fitting techniques such as previous-value, linear-curve, and quadratic-curve fitting~\cite{di2016fast}, to more complex multidimensional first-order and second-order Lorenzo predictors~\cite{tao2017significantly}, and dynamic multidimensional spline interpolation, including linear and cubic splines~\cite{zhao2021optimizing}.
HPEZ~\cite{liu2023high} introduces novel interpolation strategies (\eg  multidimensional interpolation, and natural cubic splines) along with auto-tuning driven by quality metrics. 

\stab\noindent
{\bf Neural networks for scientific data compression.}
Recently, the proliferation of deep neural network techniques has spurred the development of deep learning-based methods for scientific data compression.
There are two main categories: 
\myNum{i} The first one includes lossy methods without error bounds.
For example, Hayne et al.~\cite{hayne2021using} evaluate compression performance on 2D floating-point scientific data using an existing autoencoder~\cite{balle2018variational} designed for natural image compression.
Huang et al.~\cite{huang2023compressing} develop a coordinate-based neural network for compressing weather and climate data. Such methods report high compression ratios but the accuracy of the decompressed data may be arbitrarily low.
\myNum{ii} The second category comprises error-bounded lossy compression methods.
This includes AE-SZ ~\cite{liu2021exploring}, which incorporates a Slice-Wasserstein autoencoder network, and SRN-SZ~\cite{liu2023srn}, which employs super-resolution networks to enhance compression performance.

\stab\noindent
{\bf Remote sensing compression.}
The remote sensing community deals with scientific data, mostly in the form of images~\cite{toth2016remote}. 
Neural network-based methods for remote sensing image compression, such as COSMIC~\cite{zhang2024cosmic} and HL-RSCompNet~\cite{xiang2024remote}, are optimized for visual presentation where minor inaccuracies in pixel values are acceptable. However, they may fail to preserve the numerical precision 
required for many scientific applications; furthermore, they do not achieve competitive compression ratios\footnote{For comparison, HL-RSCompNet~\cite{xiang2024remote} on the \RS500 dataset yields  compression ratio $\rho=45.51$ {\em without error guarantee}, whereas error-bounded methods such as SPERR~\cite{li2023lossy} and SZ3.1~\cite{zhao2021optimizing} achieve $\rho=69.23$ and $\rho=77.87$, respectively, under error bound $\epsilon = 10^{-2}$.}.

\section{Graph-based compression}\label{sec:method}

In this section, we propose {\GraphCom}, a novel graph-based error-bounded lossy compression method for scientific data.
{\GraphCom}~comprises three main steps (see Figure~\ref{fig:workflow}):
\myNum{i} data segmentation and graph initialization, which partitions each timestamp into irregular regions and maps the partitioning to a graph  (Section~\ref{sec:method:graph-initialization});  
\myNum{ii} graph representation learning, which trains {\GNN}, a temporal graph autoencoder to generate latent representations of the graph (Section~\ref{sec:method:graph-representation}); and 
\myNum{iii}  grid reconstruction, which uses the trained model and the latent representations to reconstruct an approximation of the original data (Section~\ref{sec:method:graph-to-grid}). 
Similarly to existing work \cite{liu2023high,zhao2021optimizing}, {\GraphCom} is augmented by \myNum{iv} an error-bounded module, where the residual error is quantized, encoded by the Huffman algorithm and compressed losslessly (Section \ref{sec:error-bound-module}). 
This module is necessary to guarantee that the point-wise relative error of the decompressed data is bounded by a user-defined threshold $\epsilon$. 
Below, we explain these steps in detail.

\subsection{Data segmentation and graph initialization}\label{sec:method:graph-initialization}

The input to our compressor is a temporal scientific dataset
$X \in \mathbb{R}^{T\times M\times N}$. Since the data correspond to some physical process, intuitively in each timestamp there exist regions $p_i$ that are relatively smooth. For each such region, we can replace the values of all points in $p_i$ with some aggregate value, \eg the mean $\overline{p_i}$, essentially \emph{compressing} the original data. With high probability, the relative difference between the original value of each data point and $\overline{p_i}$ will be bounded by $\epsilon$. Data points not satisfying this condition, will be handled later by the error-bounded module (Section \ref{sec:error-bound-module}).      

The following questions arise:
\myNum{i} how to identify the smooth regions, which are expected to have irregular shapes;
\myNum{ii} how to represent the relationships among those regions; and
\myNum{iii} how to efficiently store the spatial extents of all regions. 

\stab
\noindent \textbf{Data segmentation.} 
To answer the first question, observe that the problem is a combination of data and space partitioning; hence, it does not map well to traditional clustering, such as $k$-means~\cite{lloyd1982least}, DBSCAN~\cite{ester1996density}, or hierarchical clustering~\cite{murtagh2012algorithms}.
Instead, the problem resembles image segmentation. Since the subsequent steps of our method require spatially connected regions, we cannot employ NN-based segmentation~\cite{minaee2021image}, such as Segnet~\cite{badrinarayanan2017segnet}, or U-Net~\cite{khalel2019multi, wang2020weakly}, because they may generate disconnected regions. Instead, we focus on segmentation methods such as Quickshift~\cite{vedaldi2008quick}, Compact Watershed~\cite{neubert2014compact} and the Felzenszwalb algorithm~\cite{felzenszwalb2004efficient}. Based on our ablation study (Section~\ref{exp:different-segmentation-algos}), we employ the latter for the rest of this paper.

The Felzenszwalb algorithm~\cite{felzenszwalb2004efficient} is a graph-based approach that treats each data point as a node with edges representing value similarities. 
The algorithm efficiently handles spatial correlations and accurately identifies natural boundaries using a minimum spanning tree. The resulting partitioning is a set of regions with irregular spatial extents; regions do not overlap and their union covers the entire space. 

\stab
\noindent \textbf{Meta-timestamp selection.}
Since data changes at each timestamp, the resulting segmentation also varies. Storing the spatial extents of the segmentation regions for every timestamp would require a lot of space which negates any compression achieved by value aggregation. Thus, we refrain from segmenting all timestamps. Instead, we carefully select a minimal yet effective set of representative timestamps (that we call \emph{meta-timestamps}) that optimize compression performance by maintaining a balance between segmentation accuracy and storage efficiency. Intuitively, an effective meta-timestamp should approximate the data distribution while minimizing temporal variations of the timestamps that it represents.

Our meta-timestamps selection is illustrated in Algorithm \ref{alg:meta-timestamps-selection}.  Interestingly, to determine the selection of meta-timestamps, we do not use the computationally expensive compression ratio (that requires the execution of the compression algorithm). We propose a more efficient evaluation that incorporates the Mean Absolute Difference (\MAbsD), defined as the mean absolute difference between the values of meta-timestamp and the mean of the group of timestamps it represents. {\MAbsD} can be efficiently computed, significantly reducing the optimization cost.

\begin{algorithm}[t]
\DontPrintSemicolon
\SetAlgoLined
\KwIn{Dataset $X \in \mathbb{R}^{T \times M \times N}$ and the maximum number of meta-timestamps $R_{\max}$}
\KwOut{Meta-timestamps set $\mathbb{M}$}

\Comment{Compute Metric matrix $D$ \& Cumulative sum $S$} 
Initialize $D, S \in \mathbb{R}^{T \times T}$  \;

\For{$t \gets 0$ \textbf{to} $T-1$}{
    $D[t,:] \gets \text{\MAbsD}(X_t, X_{0:T-1})$\;
    $S[t,:] \gets \textbf{cumsum}(D[t,:])$ \;
}

\Comment{Compute optimal meta-timestamps selection} 
Initialize $\mathit{cost} \in \mathbb{R}^{R_{\max} \times T}$ with $\infty$\;
Initialize $\mathit{path} \in \mathbb{R}^{R_{\max} \times T}$ with $-1$\;

\For{$r \gets 1$ \textbf{to} $R_{\max}$}{
    \For{$j \gets 0$ \textbf{to} $T-1$}{
        update $\mathit{cost}[r][j], \mathit{path}[r][j]$ using Eqs.~\ref{eq:update_cost_meta_selection} 
    }
}

\Comment{Backtrack to extract meta-timestamps} 
$\mathbb{M} \gets \emptyset$, $\mathit{current} \gets T-1$\;
$r^* \gets \arg\min_{r} \mathit{cost}[r][T-1]$\;
\While{$\mathit{current} \neq -1$}{
    $\text{start} \gets \mathit{path}[r^*][\mathit{current}]$\;
    $\mathbb{M}$.append $(\mathit{start}, \mathit{current})$\;
    $\text{current} \gets \mathit{start} - 1$\;
}

\Return{$\mathbb{M}$}
\caption{Meta-Timestamp Selection Algorithm}
\label{alg:meta-timestamps-selection}
\end{algorithm}

Algorithm~\ref{alg:meta-timestamps-selection} first constructs a metric matrix $D$, which captures temporal correlations across timestamps using the \MAbsD~function (Line 3). To accelerate cost evaluation, we precompute the prefix sum matrix $S$, enabling efficient range queries (Line 4). Then, a dynamic programming procedure (Lines 5-9), iteratively selects optimal meta-timestamps by updating the cost matrix \text{cost} and maintaining transition points in \text{path} using Eqs.~\ref{eq:update_cost_meta_selection}: 
\begin{equation}\label{eq:update_cost_meta_selection}
\small
    \begin{split}
        \mathit{cost}[r][j] 
        &
        = \min_{0 \leq m < j} \Big( 
        \mathit{cost}[r-1][m] 
        + \mathit{error}(D, S, m+1, j) 
        \Big)\\
        \mathit{path}[r][j] &= \arg\min_{0 \leq m < j} \Big( 
        \mathit{cost}[r-1][m] 
        \\
        &
        \quad 
        + \sum_{k=m+1}^{j} |D[r-1][k] - \mathit{mean}| 
        \Big)
        \\ 
\mathrm{where\ \ \ }        &
 \mathit{error}(D, S, m+1, j) = \sum_{k=m+1}^{j} |D[r-1][k] - \mathit{mean}|\\
\mathrm{and\ \ \ } & \mathit{mean} = \frac{S[r-1][j] - S[r-1][m]}{j - m}.
    \end{split}
\end{equation}

Finally, an efficient backtracking procedure (Lines 12-15) extracts the selected meta-timestamps, ensuring the optimal segmentation configuration. 
Algorithm \ref{alg:meta-timestamps-selection} balances compression performance and accuracy, making it scalable for large-scale datasets.

\stab \noindent
\textit{Time complexity.} 
The metric matrix $D$ and prefix sum $S$ computation (Line 1-4) requires $\calO(T^2)$ time, where $T$ is the number of timestamps. The dynamic programming (Lines 5-9) runs in $\calO(R_{\text{max}} \cdot T^2)$ time, where $R_{\text{max}}$ is the maximum number of meta-timestamps. Path backtracking (Lines 12-15) takes $\calO(R_{\text{max}} \cdot T)$ time. Overall, of the total time complexity of of Algorithm \ref{alg:meta-timestamps-selection} is $O(R_{\text{max}} \cdot T^2)$.

\stab
\noindent \textbf{Graph-based representation.}
After selecting the meta-timestamps, we apply a segmentation algorithm (\eg Felzenszwalb) to generate segmentation schemes for these timestamps. These segmentations are then propagated to neighboring timestamps, ensuring a limited number of segmentation schemes across the dataset. 
Each timestamp is then represented as a graph $G$, where nodes correspond to segmented regions, edges connect adjacent regions, and each node is assigned the mean value of its region as its feature.

\begin{definition}[Graph set $\mathcal{G}$]
Let $G = (V, E, F)$ be an undirected, unweighted graph that represents the data partitioning at a specific timestamp, where $V$ is the set of nodes and $E$ is the set of edges.
Let $A \in \{0,1\}^{|V| \times |V|}$ be the adjacency matrix of $G$.
Let $F \in \mathbb{R}^{|V|\times d}$ be a $d$-dimensional feature vector associated with each node in $G$. 
\emph{Graph set} $\mathcal{G}$ is the union of all graphs $G_{t}$ for all timestamps $t \in [0, T-1]$, \ie $\mathcal{G} = \bigcup_{t=0}^{T-1} G_t$.
\end{definition}

The mapping of the original grid data into graphs allows for the explicit representation of irregular spatial relationships. In the next section we will use it to train a temporal graph autoencoder.

\subsection{Graph representation learning}\label{sec:method:graph-representation}
In this section, we propose a temporal graph autoencoder, called {\GNN}, to encode the graph data into latent representations. 
These are compact representations of the graph data and significantly reduce the storage requirements, while preserving essential spatial and temporal information. 
For instance, temperature data may exhibit varying patterns depending on time and location. {\GNN} dynamically adapts to these variations, extracting compact latent representations that preserve critical global relationships. This adaptability ensures both high compression ratios and accurate data reconstruction, making it indispensable for downstream tasks like error-bounded compression.

\begin{figure}[t]
  \centering
  \includegraphics[width=\linewidth]{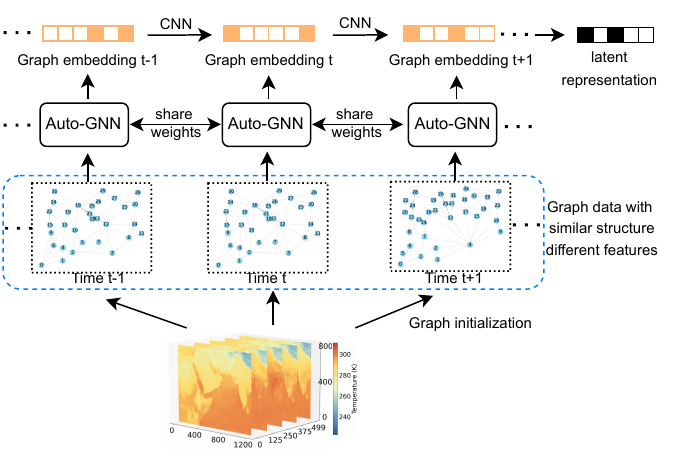}
  \caption{Overview of our temporal graph autoencoder \GNN~for latent representation, which learns spatial and temporal patterns; the decoder is omitted.}\label{fig:OverviewofAutoGNN}
\end{figure}

{\GNN} utilizes graph convolution layers to capture spatial information followed by convolution layers to capture temporal information. 
The latent representations are subsequently decoded back into graph data.
The architecture of our graph autoencoder is depicted in Figure~\ref{fig:OverviewofAutoGNN}.

\stab \noindent 
\textbf{Spatial information.} 
Building on recent advancements in graph convolution networks (GCN)~\cite{wu2020comprehensive}, we utilize graph convolution layers to learn the graph embedding. Inspired by previous work in this field~\cite{zhou2020graph}, our approach leverages the following layer-wise propagation rule in a GCN:
for each timestamp $t$, the propagation rule for the $l$-th layer uses adjacency matrix $A_t \in \{0,1\}^{|V| \times |V|}$ and graph embedding $H^{l}_t \in \mathbb{R}^{|V| \times d}$ as input, to update the graph embedding to $H^{l+1}_t$. Formally:
\begin{equation}\label{eq:graph_embedding}
\begin{aligned}
        H^{l+1}_t &= \phi\left(\tilde{A}_t H^{l}_t W^{l}\right)
\end{aligned}
\end{equation}
where $\phi$ is a non-linear activation function and $W^l \in \mathbb{R}^{d \times d}$ denotes the shared weights at layer $l$ for each timestamp.
$\tilde{A}_t$ is defined as follows (for simplicity, the timestamp index $t$ may be omitted):
\begin{equation}\label{eq:supp_graph_embedding}
        \tilde{A} = \tilde{D}^{-\frac{1}{2}} (A+I) \tilde{D}^{-\frac{1}{2}}, \quad 
        \tilde{D} = \text{diag}\left(\sum\nolimits_j \tilde{A}_{ij}\right)
\end{equation}
where $\tilde{D}$ is the degree matrix of $\tilde{A}$.
The initial graph embedding $H^{0}_t$ is derived from the graph feature $F_t$, as follows:
\begin{equation}\label{eq:initial_graph_embedding}
        H^{0}_t = F_t
\end{equation}

The final graph embedding $H^{L}_t$ is obtained through iterative application of graph convolution operations,
where $L$ denotes the final layer.

Our graph convolution enables effective propagation and transformation of node features across the graph by utilizing a normalized Laplacian to achieve smooth feature aggregation from adjacent nodes.
Consequently, \GNN~captures both local and global structures within the graph, enhancing the fidelity of the transformed features.
The graph convolution layers learn spatial information for all timestamps, each corresponding to a graph, using the shared weight matrix $W$.
The consistent structure of all graphs, resulting from the unified segmentation in graph initialization, ensures the consistent application of the learned spatial information.

The decoder transforms the encoded graph embedding $H^{L}_t$ back into an adjacency matrix through pairwise inner product calculations: $\hat{A_t} = H^{L}_t \left ( H^{L}_t \right )'$, where $\left (H^{L}_t \right )'$ denotes the transpose of the graph embedding matrix.
The decoder receives the graph embedding as input and outputs a symmetric matrix, where each entry represents the inner product between corresponding embedding. 
This transformation is crucial for accurately reconstructing graph data from their latent representations. 
Due to the effectiveness and simplicity of inner product calculations, it is extensively utilized in tasks such as graph reconstruction~\cite{kipf2016variational,wang2016structural} and is particularly advantageous for applications that require relational data representations within graph network architectures~\cite{battaglia2018relational}.

Algorithm~\ref{alg:graph-embedding} illustrates the graph embedding process, which iterates through each epoch (Line 2) and updates the model and graph embedding to minimize  loss. 
The embedding is learned through the  encoder and decoder (Lines 6-7). 
Loss is computed in Line 8 using the mean squared error (MSE) between the predicted and actual adjacency matrices. 
The weights of the model are updated using gradient descent in Lines 9-10.
After training, a graph embedding set $\mathcal{G}^L$ is initialized (Line 12) to store temporal graph embeddings (Lines 13-15).

\begin{algorithm}[t]
    \DontPrintSemicolon
    \SetAlgoLined
    \KwIn{Temporal graph set $\mathcal{G}$, epoch $E$ and learning rate $\eta$}
    \KwOut{Spatial model $\mathcal{M}$ and graph embedding $\mathcal{G}^{L}=\{H^L_t\}_{t=0}^{T-1}$}

    Initialize model \( \mathcal{M} \) and its weight $W$ \;
    \For{$\mathit{epoch} = 0$ to $E-1$}{
        Initialize \( \mathcal{L} = 0 \)  \tcp*{Loss function}
        \ForEach{$t$ in $|T|$}{
            \( F, E, A \leftarrow \mathcal{G}_i \)  \tcp*{{Node features, edges, adjacency matrix}}
    
            \( H \leftarrow \mathcal{M}.\text{encoder}(F, A, W) \) \tcp*{\small Eq.\ref{eq:graph_embedding} - Eq.\ref{eq:initial_graph_embedding}}
            \( \hat{A} \leftarrow \mathcal{M}.\text{decoder}(H, W) \) 
    
            \( \mathcal{L} \leftarrow \text{MSE}(\hat{A}, A) \)\;
            \( \Delta W \leftarrow -\eta\frac{\partial \mathcal{L}} {\partial W} \)\;
            \( W \leftarrow W + \Delta W \)\;
        }
    }
    
    \Comment{Generating graph embedding} 
    Initialize model \( \mathcal{G}^{L} = \emptyset \) \;
    \ForEach{$t$ in $|T|$}{
        \( H^L_t \leftarrow \mathcal{M}.\text{encoder}(F, A, W) \) \;
        \( \mathcal{G}^{L}.\text{add}(H^L_t) \) \;
    }
    
    \Return model $\mathcal{M}$ and graph embedding $\mathcal{G}^{L}$
 \caption{Learning Graph Embedding}\label{alg:graph-embedding}
\end{algorithm}

\stab \noindent
\textit{Time complexity.} 
The complexity of Algorithm~\ref{alg:graph-embedding} is $O(E \cdot T \cdot (|V|^2 + |W|))$.
Lines 4-10 exhibit $O(T \cdot (|V|^2 + |W|))$ complexity,
while Lines 13-15 have a complexity of $O(T)$.
The overall complexity is then determined by the outer loop over epochs (Lines 2-10), 
resulting in $O(E \cdot T \cdot (|V|^2 + |W|))$.

\stab \noindent 
\textbf{Temporal information.} 
After obtaining graph embeddings $\mathcal{G}^{L}$, we learn the latent representation through the temporal dimension $T$ to further reduce the size of graph embedding, as illustrated in Figure~\ref{fig:OverviewofAutoGNN}.
Formally, the graph embeddings are expressed as:
\begin{equation}\label{eq:all_graph_embedding}
    \mathcal{G}^{L} = \left \{H^{L}_0, \cdots, H^{L}_t, \cdots, H^{L}_{T-1} \right \}
\end{equation}
where $\mathcal{G}^{L}_T \in \mathbb{R}^{T \times |V| \times d}$.

To achieve data compression while retaining sufficient information for accurate reconstruction, we use convolutional layers to learn latent representations from the graph embeddings.
First, the convolution layers focus on temporal interactions, learning the complex relationships between nodes and their evolution over time. 
Second, these layers extract key features across temporal dimensions, effectively reducing the size of the graph embedding.
The simplified formulation is defined as follows:
\begin{equation}\label{eq:final_representation}
   \hat{\mathcal{G}}^{L}_{T-1} = g\left( f \left (\mathcal{G}^{L}_{T-1} \right ) \right )
\end{equation}
where $f(\cdot)$ denotes the encoder and $g(\cdot)$ represents the decoder.
Given the straightforward nature of the convolution process, we omit the pseudocode for brevity.

To enhance the compactness and spatial efficiency of latent representations, we implement L2 regularization. 
The loss function is formulated as:
\begin{equation}\label{eq:cnn_loss}
   \mathcal{L} = \sum_{t=0}^{T-1} \left | \left |H^{L}_t - g \left (f \left (H^{L}_t \right ) \right ) \right | \right|_2^2 + \frac{\lambda}{2} W^2
\end{equation}
where $\lambda$ is a hyperparameter, and $W$ represents the parameters of the convolutional layers.

\begin{figure}[t]
  \centering
  \includegraphics[width=.9\linewidth]{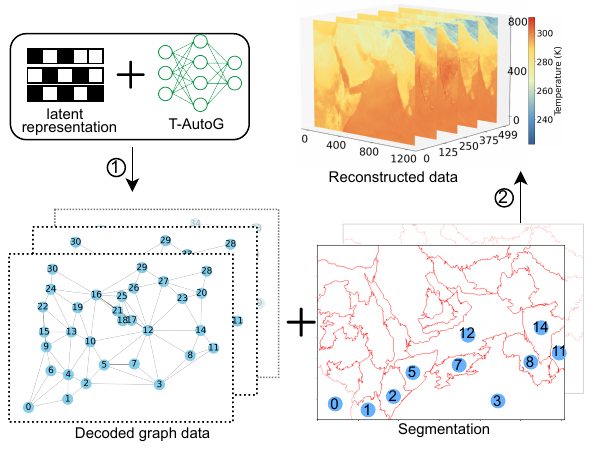}
  \caption{The overview of grid reconstruction from latent representation to reconstructed grid data.}\label{fig:grid-reconstruction}
\end{figure}

\subsection{Decompression}\label{sec:method:graph-to-grid}
After decoding the compressed representation, {\GraphCom} reconstructs the grid data by mapping decoded graph features back to their respective spatial regions using segmentations from meta-timestamps.

\stab
\noindent {\bf Grid reconstruction.}
{\GraphCom} uses {\GNN} to decode the latent representation (see  Figure~\ref{fig:grid-reconstruction}-\circlednum{1}). 
The mean value obtained from each decoded graph node is the representative temperature for the corresponding region.
Specifically, we assign the feature of each node to all the grid points within its corresponding region (as shown in Figure~\ref{fig:grid-reconstruction}-\circlednum{2}). This ensures that the reconstructed grid data accurately reflects the spatial temperature variations captured by the graph structure.

\stab
\noindent {\bf Error-bounded module.}\label{sec:error-bound-module}
In the reconstructed grid data, some values may not satisfy the relative point-wise error bound $\epsilon$. Similarly to existing research, we store the point-wise residual error in compressed form and use this information to correct the erroneous data points. 
Recall Equation~\ref{eq:formulation}: first, linear-scale quantization converts the relative difference $\frac{|x_i' - x_i|}{\mathrm{vrange}(X)}$ between each predicted $x'_i \in X'$ and original value $x_i \in X$ into integers that are multiples of $\epsilon$. 
Then, a Huffman encoder followed by Zstandard lossless compressor, store the quantized errors in compressed form.
The process is implemented in the error-bounded module (see Figure~\ref{fig:workflow}). 
For technical details, interested readers should refer to~\cite{tao2017significantly,liang2018error}.

\stab
\noindent {\bf Compression optimization.}
To further improve the compression ratio, we compress the segmentation,  the weights of {\GNN}, and the latent representation using Zstandard lossless compressor; thereby, reducing their sizes by $2\times$ and up to $5\times$.

\section{Experimental evaluation}\label{sec:experiment}
In this section, we present an experimental evaluation of {\GraphCom}.
In Section~\ref{sec:method:wgan-redsea} we also describe our implementation of a Wasserstein Generative Adversarial Network (WGAN) used to generate realistic synthetic datasets. The source code of \GraphCom~and the trained WGAN model is publicly available at \url{https://rb.gy/d2thch}.

\subsection{Experimental setup}\label{sec:exp:setting}
\stab
{\bf Software and hardware.}
\GraphCom~is implemented in {\sc Python} and {\sc PyTorch}.
It is trained on an AMD EPYC 7763 machine with 64-Core CPU@3.52GHz, 2TB RAM and NVIDIA Tesla A100 GPU with 80GB GPU memory, running Ubuntu 22.04.4 LTS (64-bit).

\stab
{\bf Datasets.}
For our evaluation, we employ the datasets presented in Table~\ref{table:redsea-data}. {\RS}  is a private dataset stemming from Red Sea reanalysis~\cite{hoteit-RSRA2018, hoteit-RSRA2022}. We use several variations that contain 500, 4K, 10K and 96K timestamps and mainly consider temperature. Still, for the largest variation (96K) we also consider humidity, and zonal and meridional wind speed (suffixed by -H, -Z and -M respectively). We also use the {\ERA}5 dataset \cite{hersbach2020era5} that contains historical whether and climate data and consider temperature and geopotential (suffixed by -T and -G respectively) for 2018 and 2023 (denoted by -18 and -23 respectively).
Moreover, we use the public datasets \Miranda, \SCALE, and {\CESM} that contain turbulence, weather and climate simulation data respectively \cite{zhao2021optimizing} (the number of variables of each dataset appear in parenthesis). Finally, use the synthetic {\WGAN} dataset that is produced by a WGAN model\footnote{Please see Section~\ref{sec:method:wgan-redsea} for details.} that generates synthetic data that closely resemble {\RS} (thus there is a one-to-one correspondence between {\WGAN} and {\RS} characteristics). Since {\RS} is private, we believe that the publicly available {\WGAN} increases reproducibility and offers the ability to other researchers to test their methods on a very large dataset.

\begin{table}[t]
\centering
\caption{Datasets for evaluation}\label{table:redsea-data}
\resizebox{\linewidth}{!}{
\begin{tabular}{clcrcc}
\hline
& \bf Dataset & \bf Dimensions  &  \bf Size\ \   & \bf \multirow{2}{*}{Measuring} & \bf Repr. \\
& \bf name    & $T \times M \times N$ & \bf in GB& & \bf in bits\\
\hline
\hline
\multirow{7}{*}{\begin{turn}{90}Real (private)\end{turn}}
& \RS500 & $500\times 855\times 1,215$   & 1.9    & Temperature & 32 \\
& \RS4K & $4,000\times 855\times 1,215$   & 15.5    & Temperature & 32 \\
& \RS10K & $10,000\times 855\times 1,215$   & 38.7    & Temperature & 32 \\
& \RS96K & $96,407\times 855\times 1,215$   & 373.1    & Temperature & 32 \\
& \RS96K-H &  $96,407\times 855\times 1,215$  &   373.1 & Humidity  &  32  \\
& \RS96K-Z & $96,407\times 855\times 1,215$   & 373.1 & Zonal wind speed & 32 \\
& \RS96K-M & $96,407\times 855\times 1,215$   & 373.1 & Mer. wind speed & 32 \\ 
\hline
\multirow{7}{*}{\begin{turn}{90}Real (public)\end{turn}}
& \ERA5-18-T & $8,760\times 721\times 1,440$   & $67.8$    & Temperature & 64 \\
& \ERA5-23-T & $8,760\times 721\times 1,440$   & $67.8$    & Temperature & 64 \\
& \ERA5-18-G & $8,760\times 721\times 1,440$   & $67.8$    & Geopotential & 64 \\
& \ERA5-23-G & $8,760\times 721\times 1,440$   & $67.8$    & Geopotential & 64 \\
& \Miranda  &   $256\times 384\times 384$  &   1  &   Turbulence (7) &  32  \\ 
& \SCALE  &  $98\times 1,200\times 1,200$  &   6.4  &   Climate (12)  &  32  \\ 
& \CESM  &  $26\times 1,800\times 3,600$  &   17  &   Weather (32) &  32  \\ 
\hline
\multirow{4}{*}{\begin{turn}{90}Synthetic\end{turn}}
& \WGAN500 & $500\times 855\times 1,215$   & 1.9    & Temperature & 32 \\
& \WGAN4K & $4,000\times 855\times 1,215$   & 15.5    & Temperature & 32 \\
& \WGAN10K & $10,000\times 855\times 1,215$   & 38.7    & Temperature & 32 \\
& \WGAN96K & $96,407\times 855\times 1,215$   & 373.1    & Temperature & 32 \\
\hline
\end{tabular}}
\end{table}

\stab
{\bf Benchmarked methods.}
We compare {\GraphCom} against the following error-bounded lossy compression methods:

\begin{LaTeXdescription}
\item[SPERR~\cite{li2023lossy}]
that is based on the  SPECK wavelet algorithm~\cite{pearlman2004efficient,tang2006three}; it supports parallel processing and error tolerance.

\item[ZFP~\cite{lindstrom2014fixed}]
that implements a lifted, orthogonal block transform and embedded coding, enabling truncation of each per-block bit stream.

\item[SZ3.1~\cite{zhao2021optimizing}]
that uses a dynamic spline interpolation approach, enhanced by various optimization strategies, to increase data prediction accuracy and improve compression.

\item[HPEZ~\cite{liu2023high}]
that includes novel interpolation methods (\eg interpolation re-ordering, multi-dimensional interpolation, and natural cubic splines) and auto-tuning driven by quality metrics to enhance compression. 
\end{LaTeXdescription}  

We omit AE-SZ~\cite{liu2021exploring} and SRN-SZ~\cite{liu2023srn} since HPEZ and SZ3.1 represent the state-of-the-art amongst all SZ-based works. We have also evaluated Gorilla \cite{pelkonen2015gorilla} and HL-RSCompNet~\cite{xiang2024remote}. {\GraphCom} outperforms these methods by more than $10\times$ (and up to $100\times)$. Thus, we do not include these results in our evaluation. 

\stab
{\bf Default parameters.}
The default parameters in {\GraphCom} are assigned as follows. 
For graph initialization, the Felzenszwalb segmentation algorithm is configured with  
$\langle \text{scale}, \sigma, \text{minSize} \rangle = \langle 10, 1, 1 \rangle$.  
\begin{figure}[t]
        \centering
        \includegraphics[width=.5\linewidth]{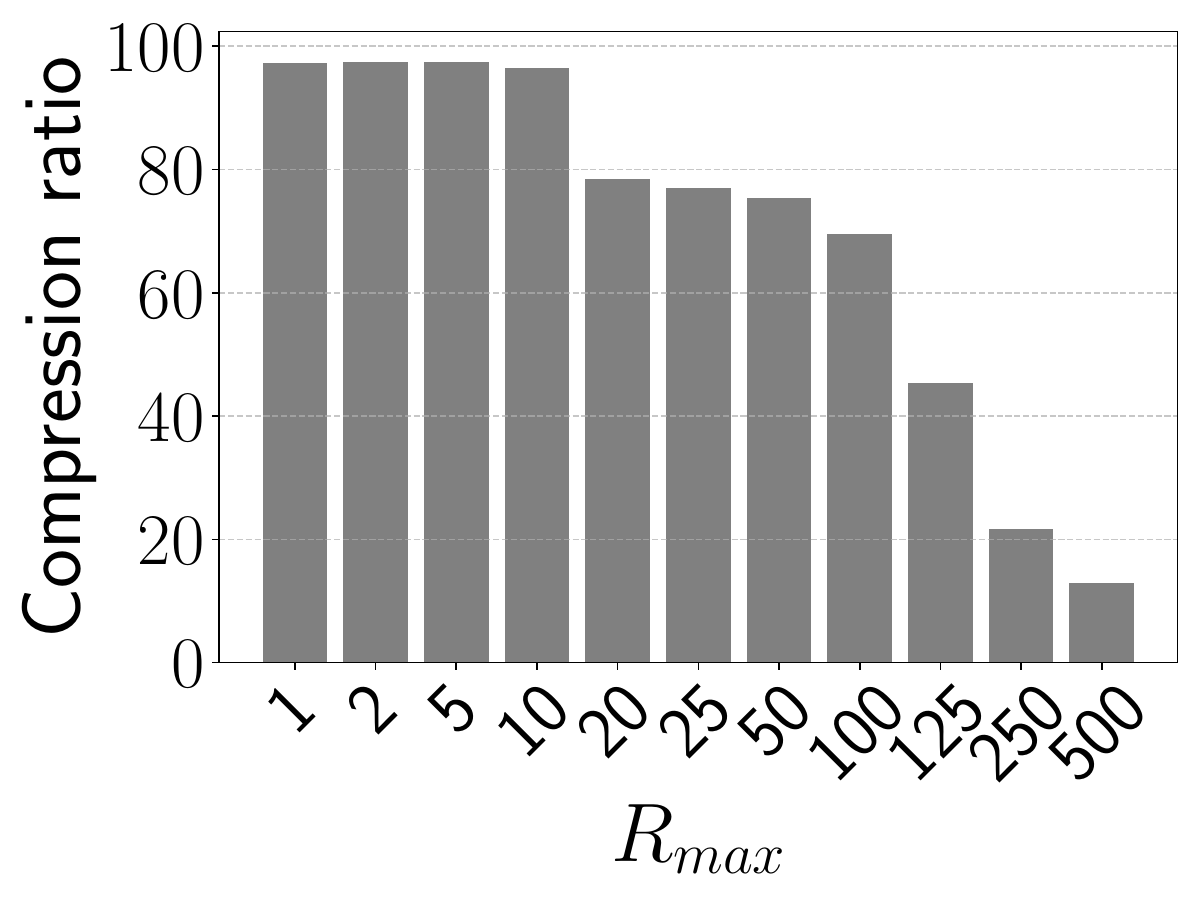}
        \label{fig:cr-different-timestamps-multiple}
    \caption{Compression ratios for the \RS500 dataset for different $R_{\max}$ values.}
    \label{fig:cr-different-timestamps}
\end{figure}

To determine the maximum number of meta-timestamps, $R_{\max}$ we evaluate the compression ratio of the \RS500 dataset for several choices of $R_{\max}$. Our results are presented in Figure~\ref{fig:cr-different-timestamps}. 
We may see that increasing $R_{\max}$ initially improves the compression ratio. However, beyond a certain threshold (approximately 10), a further increase degrades the compression ratio. Thus, we set $R_{\max} = 10$.

For \GNN~architecture, the model integrates a graph neural network (GNN) for spatial dependencies and a convolutional neural network (CNN) for temporal modeling.  
The GNN employs a three-layer graph convolution network (GCN) with hidden dimensions of 64, 128, and 256, using ReLU activation.  
The CNN processes sequences of graph embeddings with three convolutional layers, an input channel of 256, output channels of 64, 128, and 256, and kernel sizes ranging from 3 to 5.  
The network uses a default batch size of 16, adjustable based on GPU memory. The learning rate is tuned across $10^{-3}$, $10^{-4}$, and $10^{-5}$ to optimize performance. 


\begin{table}[t]
\caption{Compression ratio $\rho$ for various datasets and error bounds. \textbf{Bold} values indicate best performance; \underline{underlined} values are runner-ups.}\label{table:final-cr-real-gan}
\resizebox{\linewidth}{!}{
\begin{tabular}{lcrrrrr}
\hline
\bf  Dataset & $\epsilon$ & \bf SPERR & \bf ZFP  & \bf SZ3.1 & \bf HPEZ  & \GraphCom \\
\hline\hline
        & $10^{-2}$ & 92.54 & 8.53 & \underline{105.21} & 102.59&  {\bf 141.75}  \\
\RS96K      & $10^{-3}$ & 14.20  & 4.15 & 14.94  & \underline{15.29} &  {\bf 19.14}   \\
              & $10^{-4}$ & 5.85  & 2.99 & \underline{6.13}   & 5.95  &  {\bf 7.65}    \\ \hline
                & $10^{-2}$ & 61.86 & 5.97 & \underline{79.63}  & 74.06 &  {\bf 100.12}    \\
\RS96K-H & $10^{-3}$ & 11.28 & 3.83 & 12.70   & \underline{12.77} &  {\bf 15.73}   \\
              & $10^{-4}$ & 5.28  & 2.59 & 5.54   & \underline{5.64}  &  {\bf 6.93}    \\ \hline
                & $10^{-2}$  & 63.39 & 6.70 & \underline{65.70}  & 63.99 &  {\bf 88.66}   \\
\RS96K-Z & $10^{-3}$ & 11.35 & 4.12 & 11.44  & \underline{12.00}    &  {\bf 14.84}    \\
              & $10^{-4}$ & 5.07  & 2.72 & 5.22   & \underline{5.45}  &  {\bf 6.88}    \\ \hline
                & $10^{-2}$  & 55.77 & 6.63 & \underline{79.24}  & 74.41 &  {\bf 103.59}   \\
\guozhong{\RS96K-M}  & $10^{-3}$ & 10.68 & 4.09 & 12.48  & \underline{13.10} &  {\bf 15.93}   \\
              & $10^{-4}$ & 5.22  & 2.92 & 5.45   & \underline{5.63}  &  {\bf 6.98}    \\ \hline
                & $10^{-2}$  & 219.81& 9.98 & \underline{370.19} & 293.15&  {\bf 483.21}  \\
\ERA5-18-T    & $10^{-3}$ & 30.16 & 5.41 & 32.29  & \underline{33.39} &  {\bf 45.34}   \\
              & $10^{-4}$ & 9.52  & 3.59 & 9.18   & \underline{9.71}  &  {\bf 14.15}   \\ \hline
                & $10^{-2}$  & 221.15& 9.95 & \underline{374.94} &295.06 &  {\bf 491.27}  \\
\ERA5-23-T    & $10^{-3}$ & 30.31 & 5.39 & 32.46 &  \underline{32.59} &  {\bf 47.62}   \\
              & $10^{-4}$ & 9.55  & 3.59 & 9.24  &  \underline{9.56}  &  {\bf 14.29}   \\ \hline 
                & $10^{-2}$  & 569.38& 13.19&\underline{858.46} & 849.84 &  {\bf 1,107.57} \\
\ERA5-18-G    & $10^{-3}$ & 56.30  & 7.09 & 79.95  & \underline{80.01} &  {\bf 106.72}   \\
              & $10^{-4}$ & 17.28 & 4.29 & 15.91  & \underline{18.49} &  {\bf 25.29}   \\ \hline
                & $10^{-2}$  & 572.89& 13.8 & \underline{860.51} & 852.26& {\bf 1,130.25}  \\
\ERA5-23-G    & $10^{-3}$ & 55.19 & 7.29 & \underline{80.23} &  80.12 & {\bf 104.94}    \\
              & $10^{-4}$ & 17.71 & 4.38 & 15.92 &  \underline{19.65} & {\bf 26.42}    \\ \hline
              & $10^{-2}$  & 971.40& 46.60 & 574.60 & {\bf 1320.00} & \underline{998.82}  \\
\Miranda       & $10^{-3}$ & 243.90 & 25.60 & 168.00 & {\bf 258.00} & \underline{245.42}    \\
              & $10^{-4}$ & {\bf 74.50} & 14.50 & 47.30 &  63.60 & \underline{66.71}    \\ \hline
              & $10^{-2}$  & 103.50 & 14.50 & \underline{167.30} & {\bf 186.00} & 147.36  \\
\SCALE   & $10^{-3}$ & 35.30 & 7.80 & \underline{40.40} & {\bf 52.90} & 39.75    \\
              & $10^{-4}$ & \underline{15.00} & 4.60 & 14.10 &  {\bf 15.40} & 14.92    \\ \hline              
              & $10^{-2}$  & {\bf 1221.00} & 18.20 & 373.00 & \underline{675.00} & 497.69  \\
\CESM      & $10^{-3}$ & \underline{150.00} & 9.60 & 64.90 & {\bf 153.00} & 81.47    \\
              & $10^{-4}$ & \underline{35.00} & 5.80 & 22.90 &  {\bf 38.90} & 24.60    \\ 
\hline
\end{tabular}}
\end{table}

\subsection{Overall compression ratio}\label{exp:results-ovreall-cr}
Table~\ref{table:final-cr-real-gan} presents the compression ratio $\rho$ of all real large datasets, for three relative error bounds, $\epsilon = 10^{-2}, 10^{-3}, 10^{-4}$. It is clear that {\GraphCom} outperforms its competitors in most tested cases, achieving a 22\% to 50\% improvement. For example, for $\epsilon = 10^{-2}$ {\GraphCom} compresses \RS96K (\ie temperature data) by $141.75\times$, meaning that the original size of 373.1GB is reduced to 2.63GB. For the same settings, the next best method, SZ3.1, achieves $\rho = 105.21$; therefore, {\GraphCom} is 35\% better.

As expected, for tighter error bounds (\eg $\epsilon = 10^{-4}$) the compression ratio of all methods decreases. When the error bound is tighter, the probability of erroneous predictions increases for all methods. Consequently, the corresponding error-bounding modules must store additional information to correct the residual error, thus decreasing the final compression ratio.

Observe that different variables in the same dataset behave differently. For example, \RS96K represents atmospheric temperatures, whereas \RS96K-Z is the zonal wind speed (\ie the projection of wind speed vector to the $x$-axis). For the former, all methods achieve better compression because temperature data are smoother and easier to predict. 
Also note that all \ERA5 data are more compressible than the {\RS} data since the spatial resolution of \ERA5 is lower, meaning that each data point corresponds to an aggregation of measurements of a larger region on the Earth's surface; therefore, the distribution of values tends to be smoother.  

\begin{figure}[t]
    \centering
    \begin{subfigure}[b]{0.89\linewidth}
        \centering
        \includegraphics[width=.8\linewidth]{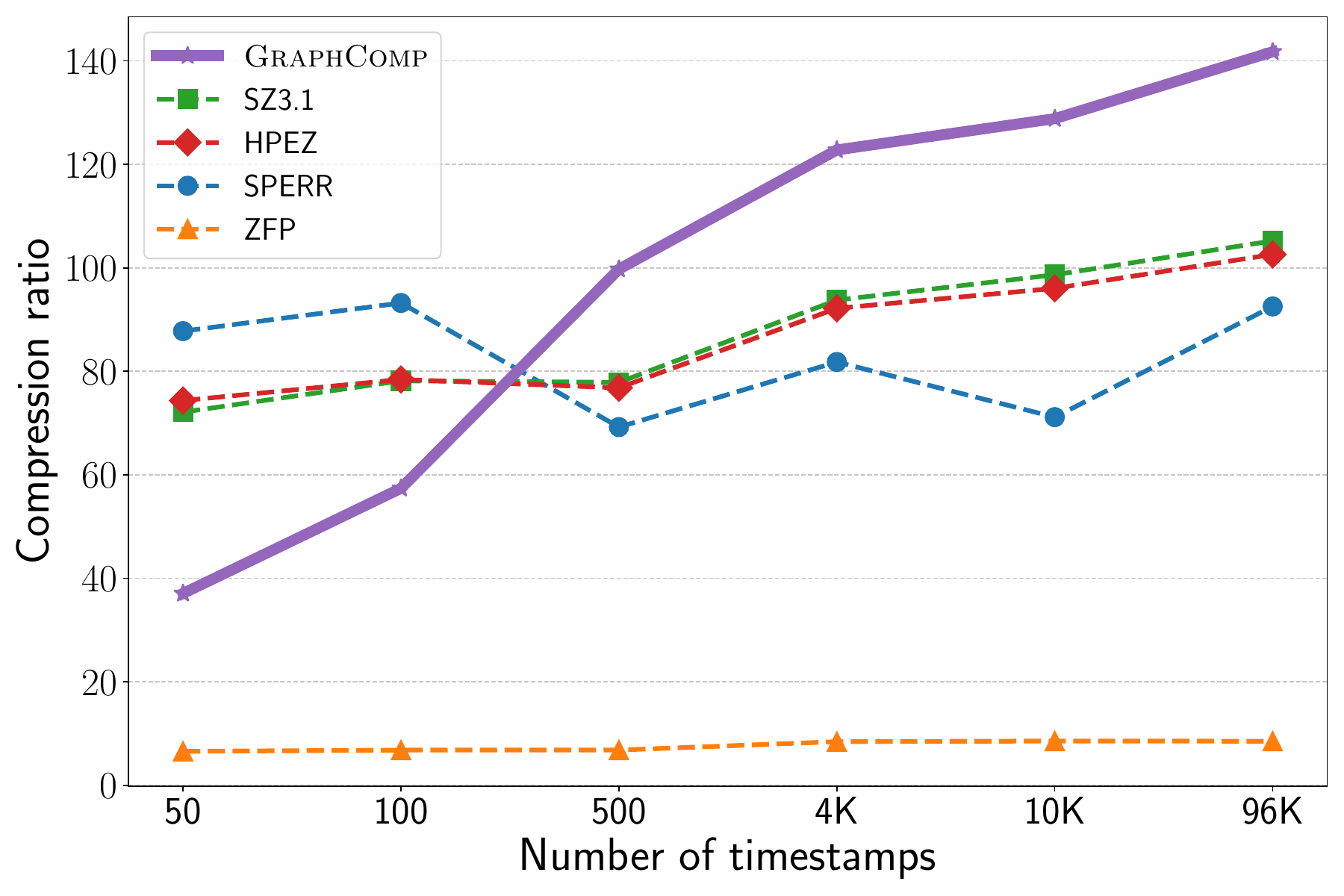}
    \end{subfigure}%
    \caption{Compression ratio of the {\RS}96K dataset using a different number of timestamps ($\epsilon = 10^{-2}$).}
    \label{fig:cr-varying-timestamps}
\end{figure}

{\GraphCom} outperforms all methods for all dataset with a large number of timestamps (\RS~and \ERA5). For the smaller-sized datasets (\Miranda, \SCALE~and \CESM) than contain less than 500 timestamps the performance of {\GraphCom} is not the best. This can be attributed to the fact that a limited number of timestamps restrict {\GraphCom} ability to exploit temporal correlations, thus, offering a lower compression ratio. To verify our claim, we use the {\RS} dataset and compute the compression ratio   $\rho$ for  different numbers of timestamps ranging from 50 to 96K. The results of Figure~\ref{fig:cr-varying-timestamps}  show that {\GraphCom} achieves substantial improvements in $\rho$ with increasing number of timestamps, while methods like HPEZ and SZ3.1 show only minimal gains. These results demonstrate \GraphCom's ability to scale and effectively utilize temporal correlations for compression. Additionally, these results suggest that increasing the number of timestamps of smaller sized datasets such as \Miranda, \SCALE, and \CESM~ would similarly enhance \GraphCom’s performance, potentially outperforming other methods.

\begin{figure}[t]
    \centering
    \begin{subfigure}[b]{\linewidth}
        \centering
        \includegraphics[width=\linewidth]{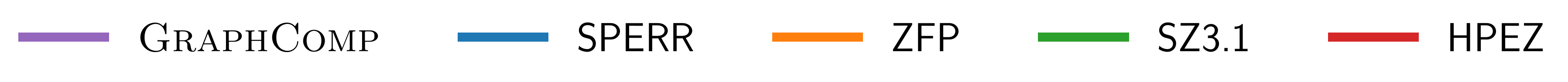}
    \end{subfigure}
    \begin{subfigure}[b]{0.48\linewidth}
        \centering
        \includegraphics[width=\linewidth,height=3cm]{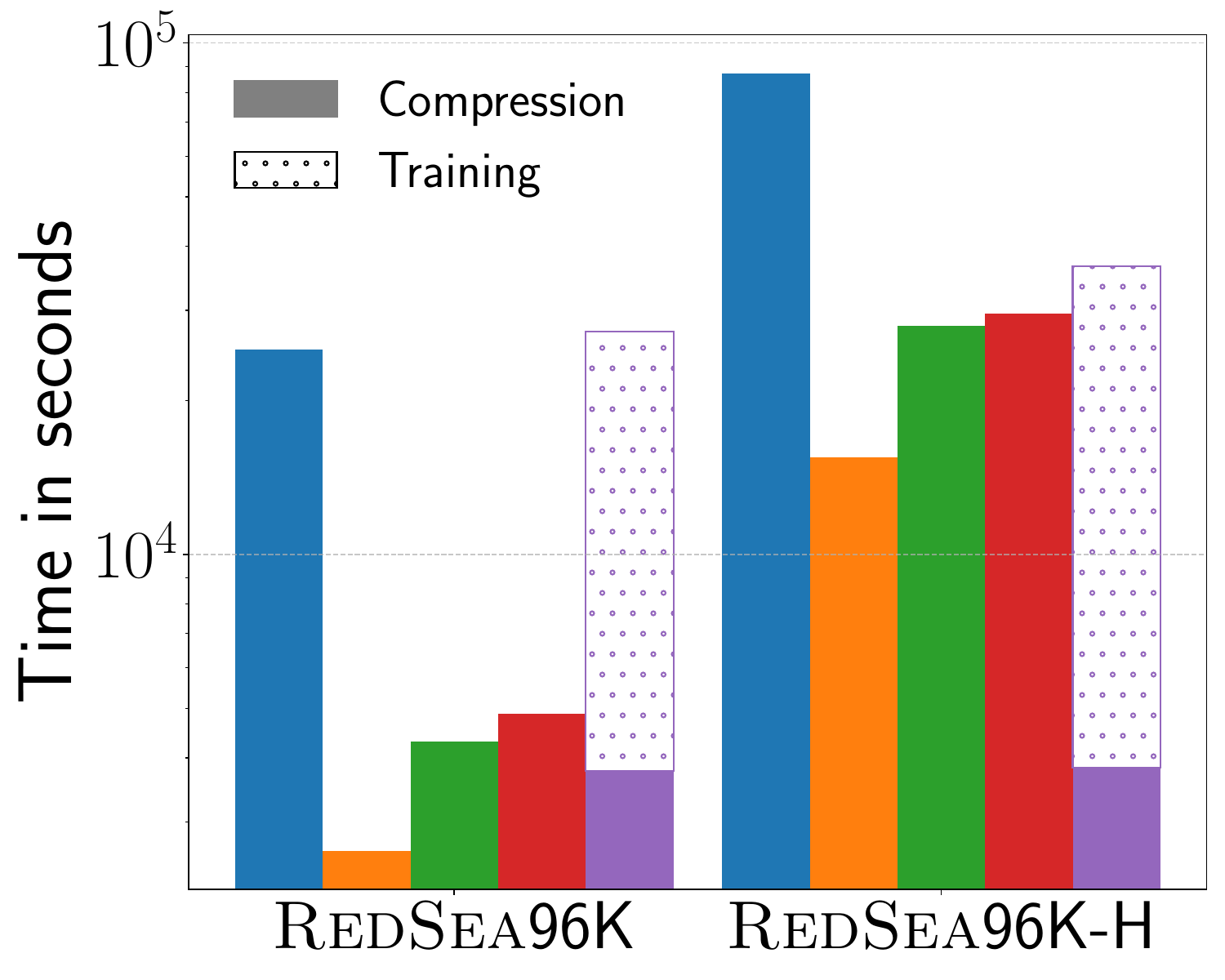}
        \caption{Compression and Training}
        \label{fig:efficiency-compression-training}
    \end{subfigure}
    \begin{subfigure}[b]{0.48\linewidth}
        \centering
        \includegraphics[width=\linewidth,height=3cm]{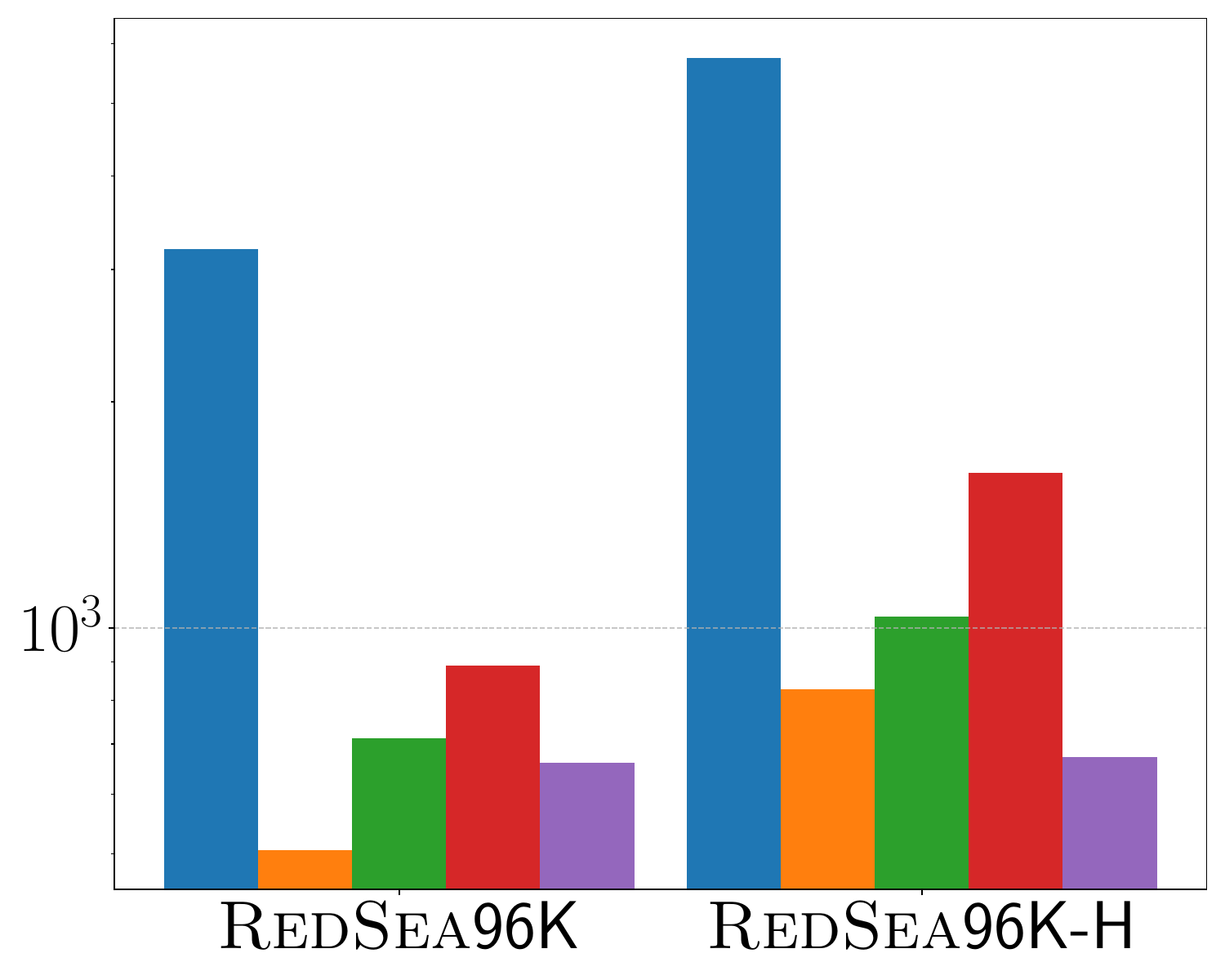}
        \caption{Decompression}
        \label{fig:efficiency-decompression}
    \end{subfigure}
    \caption{Time of compression, training, and decompression for the compared methods ($\epsilon = 10^{-2}$).}
    \label{fig:exp:compression-decompression-efficiency}
\end{figure}

\subsection{Overall efficiency}
Figure~\ref{fig:exp:compression-decompression-efficiency} illustrates the compression and decompression time (in seconds) for the compared methods presented on a logarithmic scale. \GraphCom’s compression and decompression times remain competitive with ZFP and SZ3.1, two of the fastest methods, highlighting its efficiency in these operations.

{\GraphCom} has an additional cost for training (also shown in Figure~\ref{fig:exp:compression-decompression-efficiency}). Still, this cost may be amortize for several compression tasks. Consider for instance the scenario that datasets are produced in a site and distributed between several peers. In this case, we may  train the model once, share it between all peers and use it whenever data compression, exchange and decompression is required. 

\begin{figure}[t]
    \centering
    \begin{subfigure}[b]{\linewidth}
        \centering
        \includegraphics[width=\linewidth]{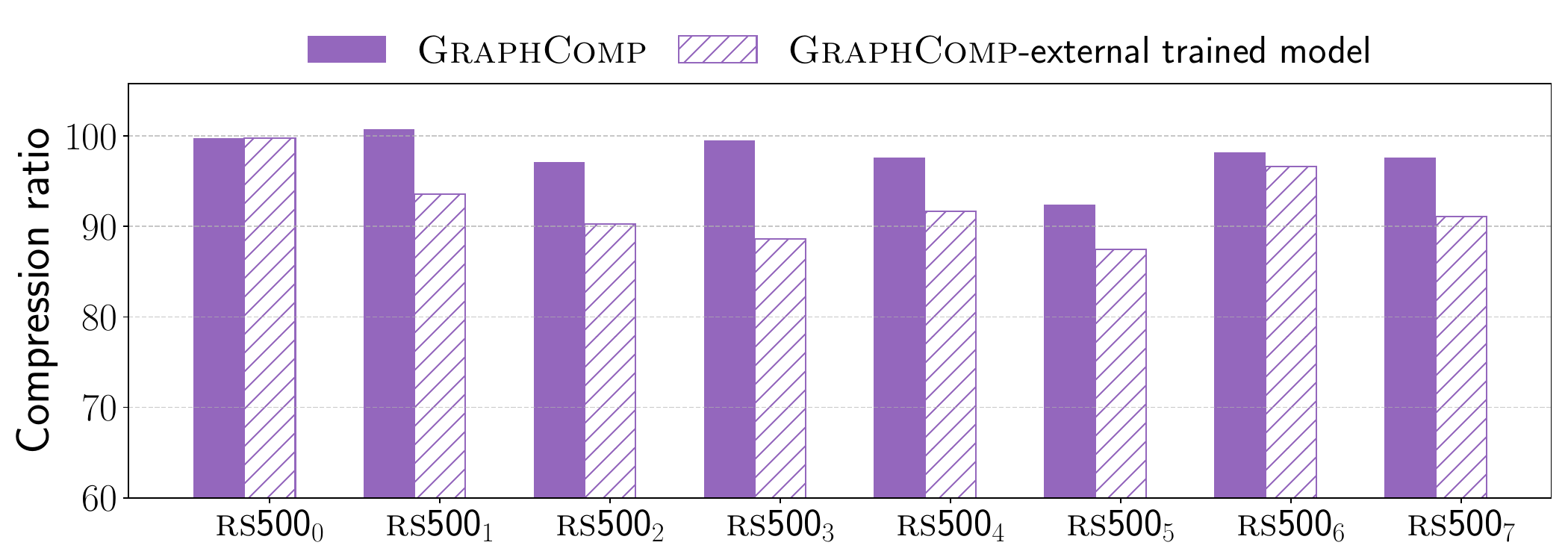}
    \end{subfigure}%
    \caption{Comparing the compression ratio of standard and external trained {\GraphCom} ($\epsilon = 10^{-2}$)}
    \label{fig:exp:cr-model-based-redsea500}
\end{figure}

To test the reusability of the model, we perform the following experiment. We consider the \RS96K dataset and construct datasets $\textsc{rs}500_{i}$, $i\in[0,7]$, each containing the 500 timestamps with indexes in the $[500i,\; 500 (i+1)]$ range of the \RS96K dataset.

We train the compression model for the $\textsc{rs}500_{0}$ dataset and call it the \emph{external trained model}. Then, consider all datasets $\textsc{rs}500_{0}$ to $\textsc{rs}500_{7}$ and compress them \emph{(a)} using standard {\GraphCom} (\ie the training is performed with the input data) and \emph{(b)} using a variation of {\GraphCom} that does not perform training but instead uses the external trained model of $\textsc{rs}500_{0}$ (\ie no training is performed). We illustrate the results in Figure \ref{fig:exp:cr-model-based-redsea500}. In all cases external training achieves competitive results, demonstrating the effectiveness of the general-purpose model for efficient compression without training costs.

\subsection{Rate distortion analysis}

\begin{figure}[t]
    \centering
    \begin{subfigure}[b]{.9\linewidth}
        \centering
        \includegraphics[width=\linewidth]{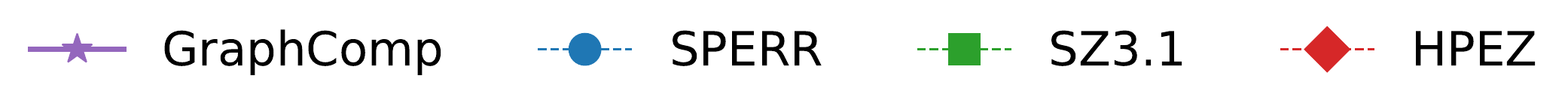}
    \end{subfigure}
    \begin{subfigure}[b]{0.49\linewidth}
        \centering
        \includegraphics[width=\linewidth]{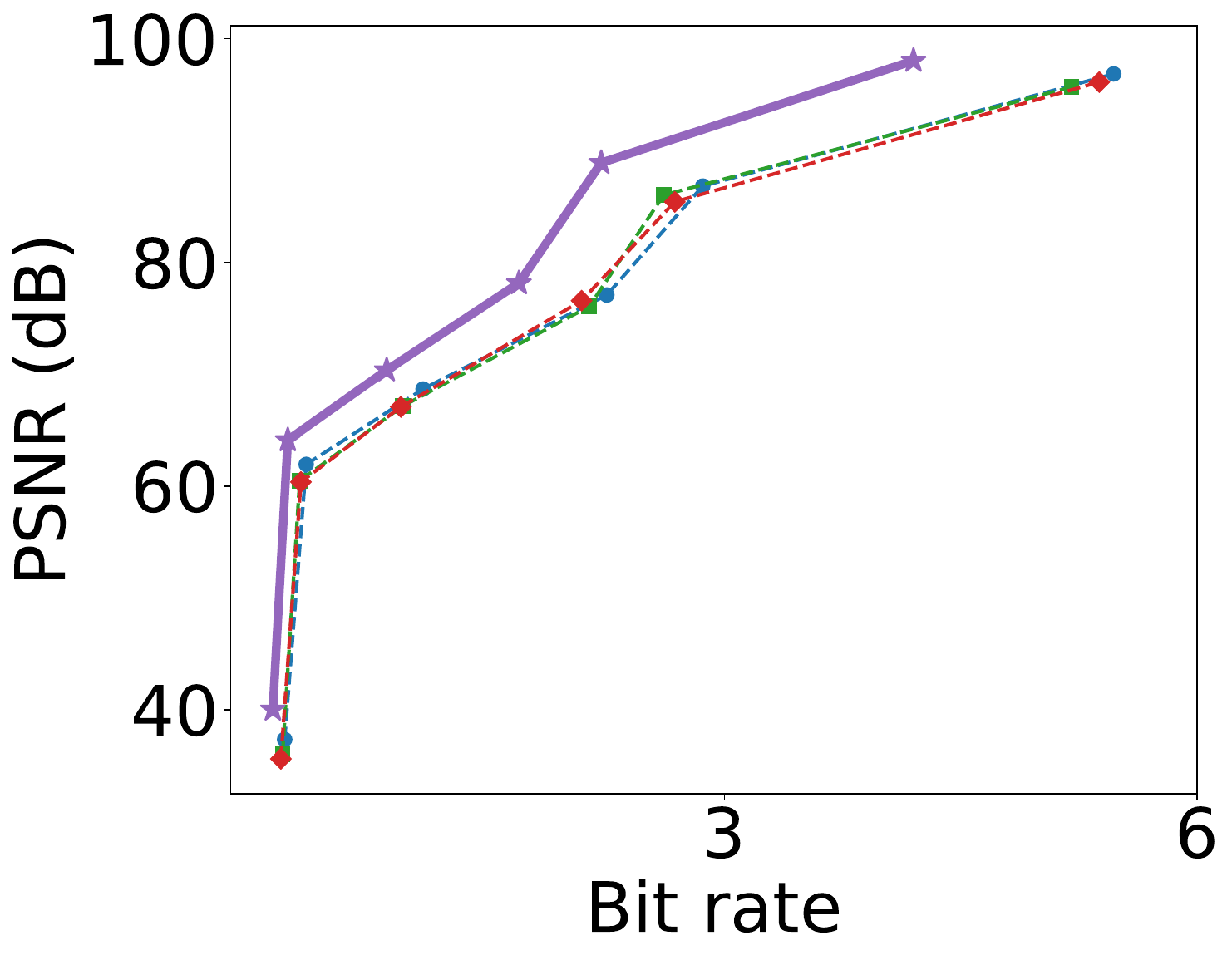}
        \caption{\RS96K}
    \end{subfigure}%
    \begin{subfigure}[b]{0.49\linewidth}
        \centering
        \includegraphics[width=.98\linewidth]{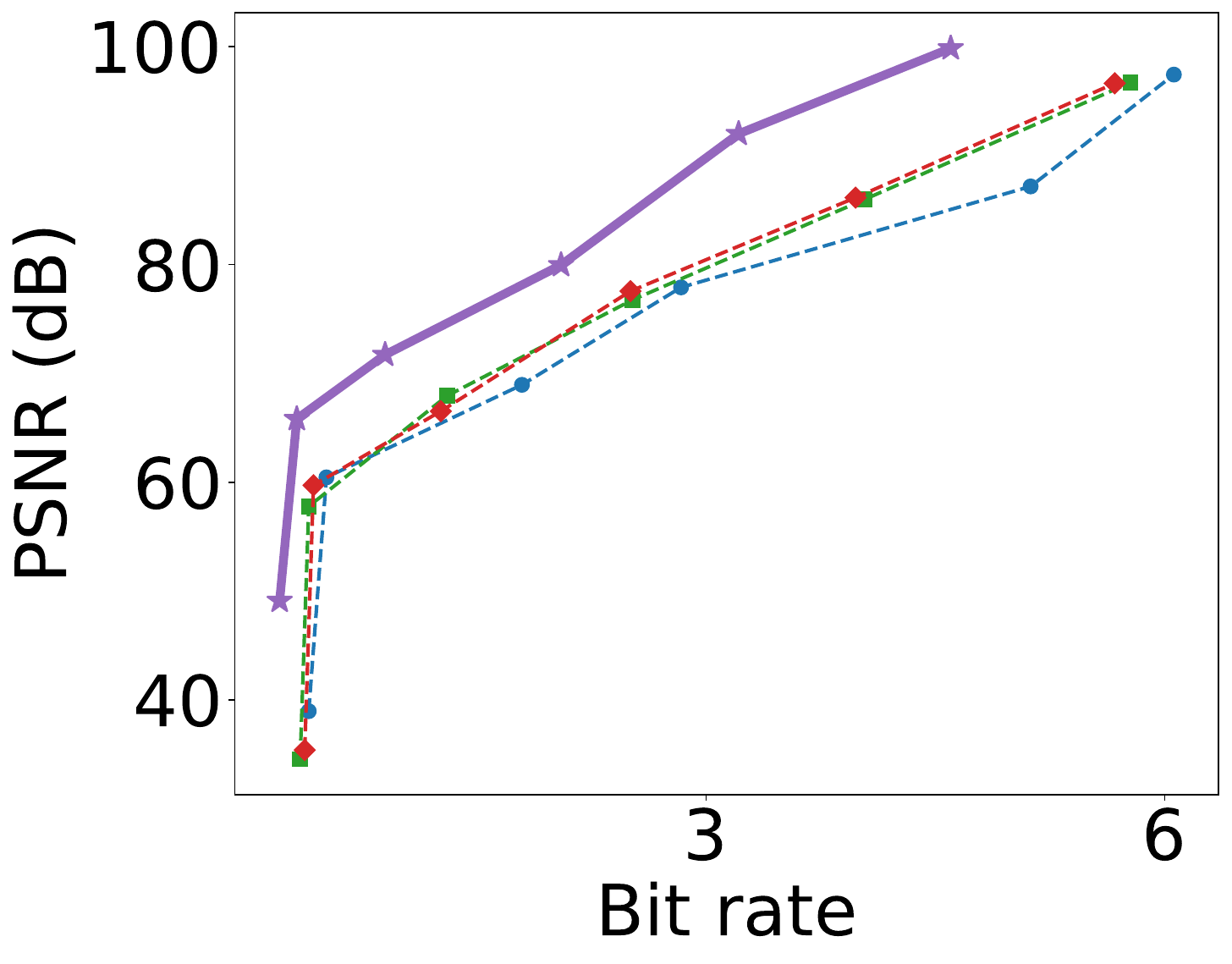}
        \caption{\RS96K-H}
    \end{subfigure}%
    \hfill
    \begin{subfigure}[b]{0.49\linewidth}
        \centering
        \includegraphics[width=\linewidth]{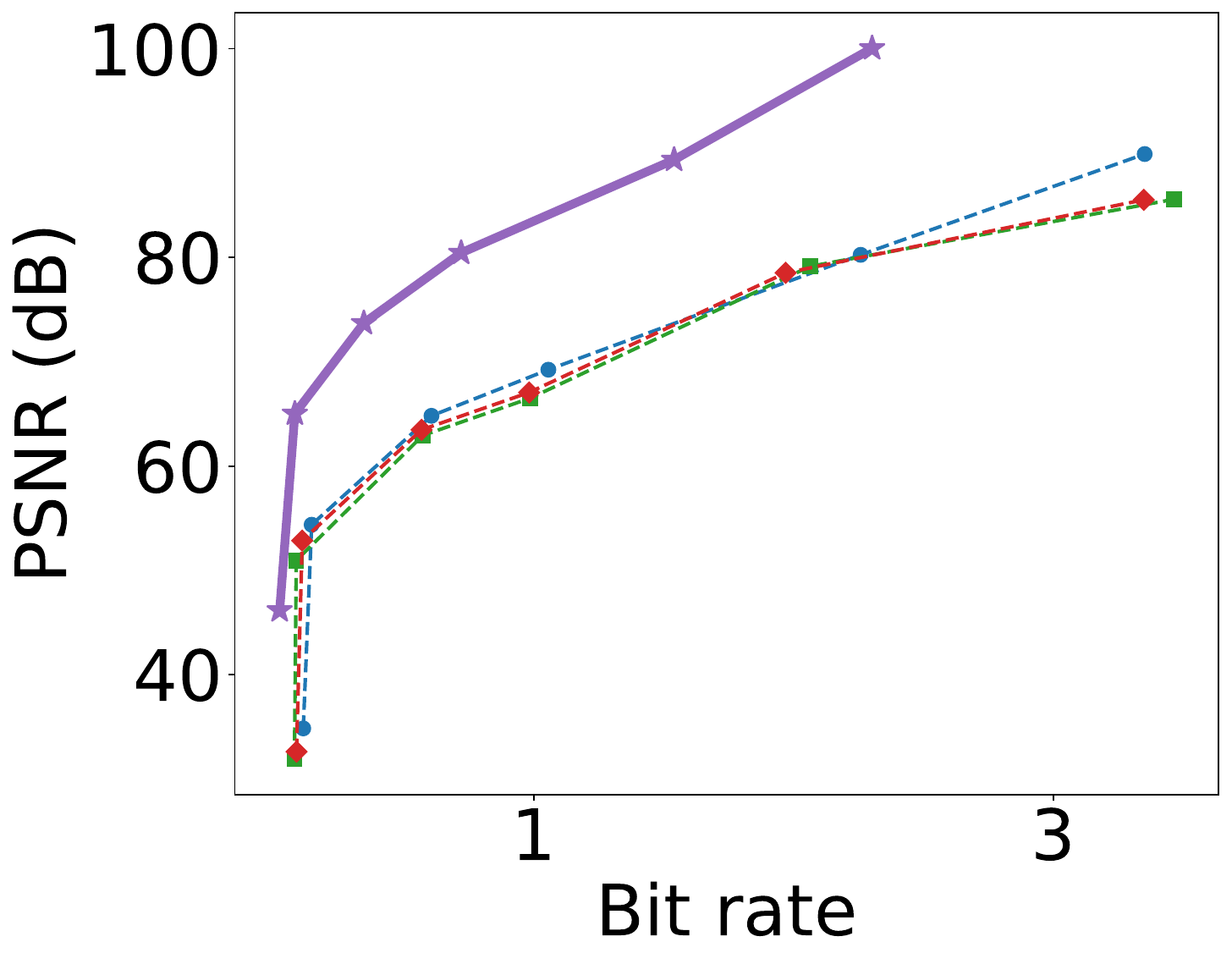}
        \caption{\ERA5-23-T}
    \end{subfigure}%
    \begin{subfigure}[b]{0.49\linewidth}
        \centering
        \includegraphics[width=.96\linewidth]{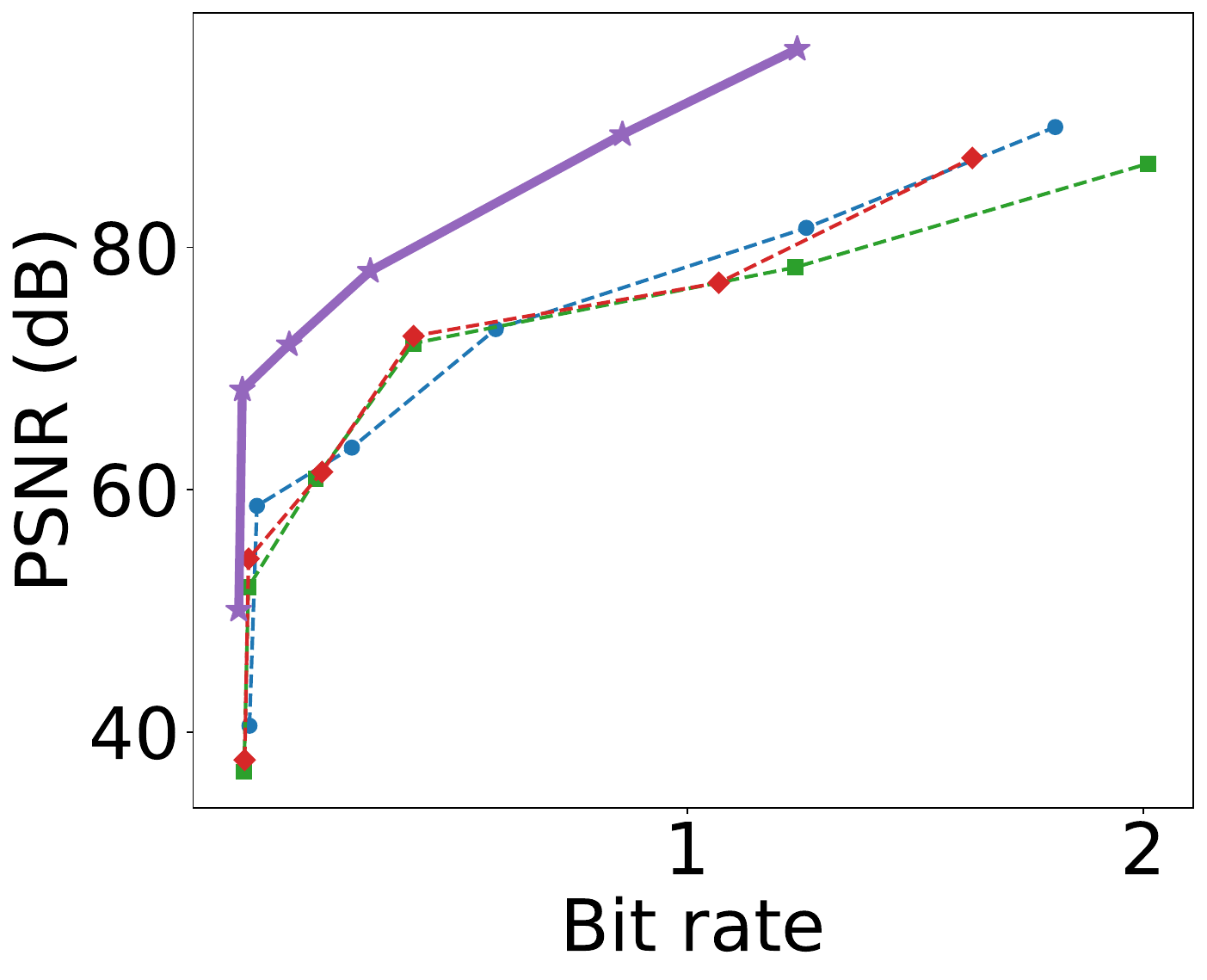}
        \caption{\ERA5-23-G}
    \end{subfigure}%
    \caption{Decompression quality (PSNR -- larger is better) vs. efficiency (bit rate -- lower is better).}
     \label{fig:exp:psnr-analysis}
\end{figure}

\noindent
{\bf PSNR analysis.} 
We evaluate the decompression quality using PSNR~\cite{zhao2021optimizing} (larger values are better) with respect to compression efficiency measured by bit rate (lower value are better). Our results are illustrated in  Figure~\ref{fig:exp:psnr-analysis}.

{\GraphCom} consistently outperforms its competitors by offering higher PSNR values at similar or lower bit rates, effectively balancing data quality and compression. 
For instance, in datasets such as {\RS} and \ERA5-23-T, \GraphCom\ achieves up to 50\% lower bit rate than the closest competitor at comparable levels of data distortion. 
For the clarity of the illustrated graphs ZFP is not presented, since it low compression ratio and high bit rate rendering graphs hard to read. 

\begin{figure}[t]
    \centering
    \begin{subfigure}[b]{0.49\linewidth}
        \centering
        \includegraphics[width=\linewidth]{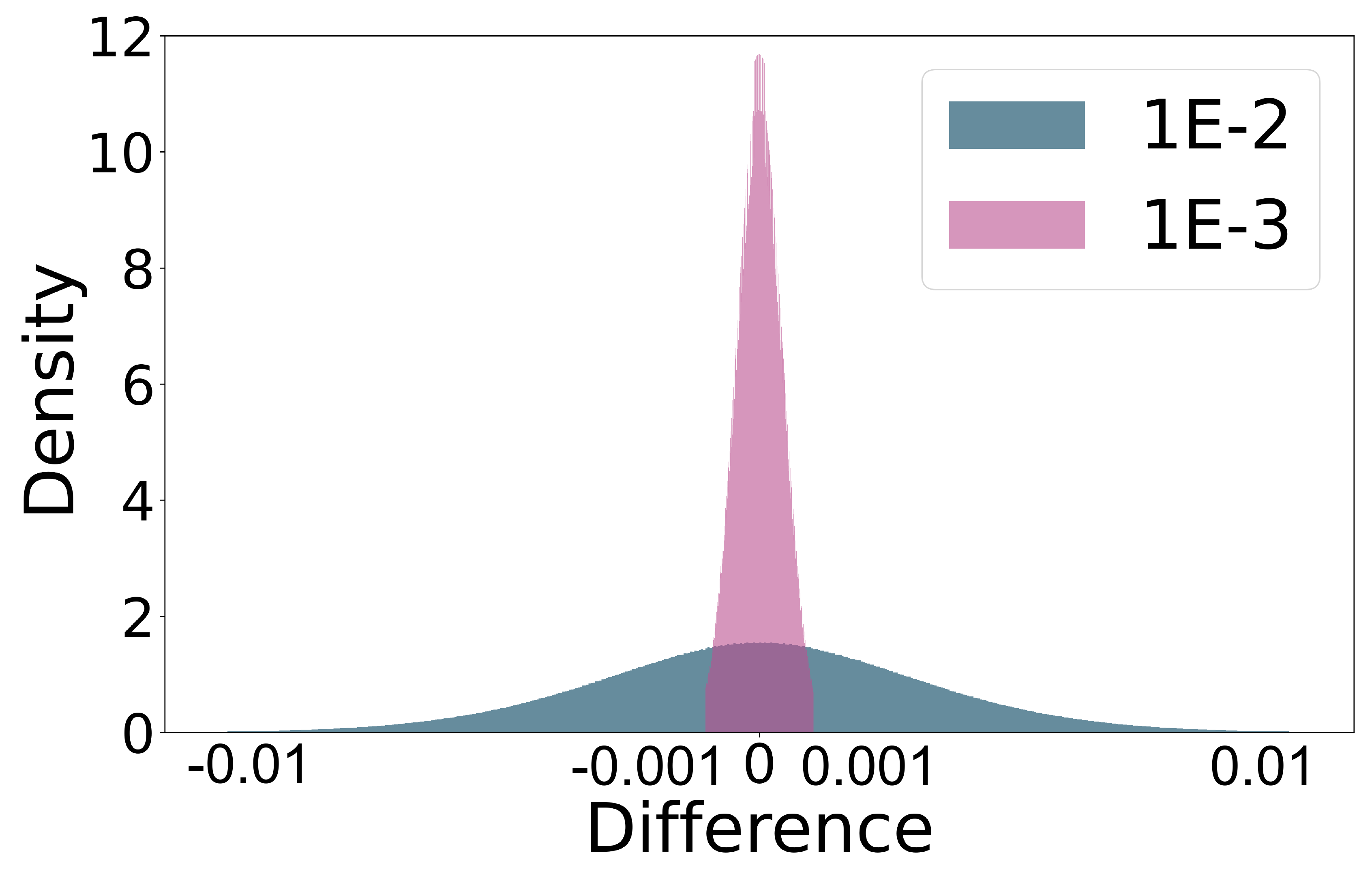}
        \caption{SPERR}
    \end{subfigure}%
    \begin{subfigure}[b]{0.49\linewidth}
        \centering
        \includegraphics[width=\linewidth]{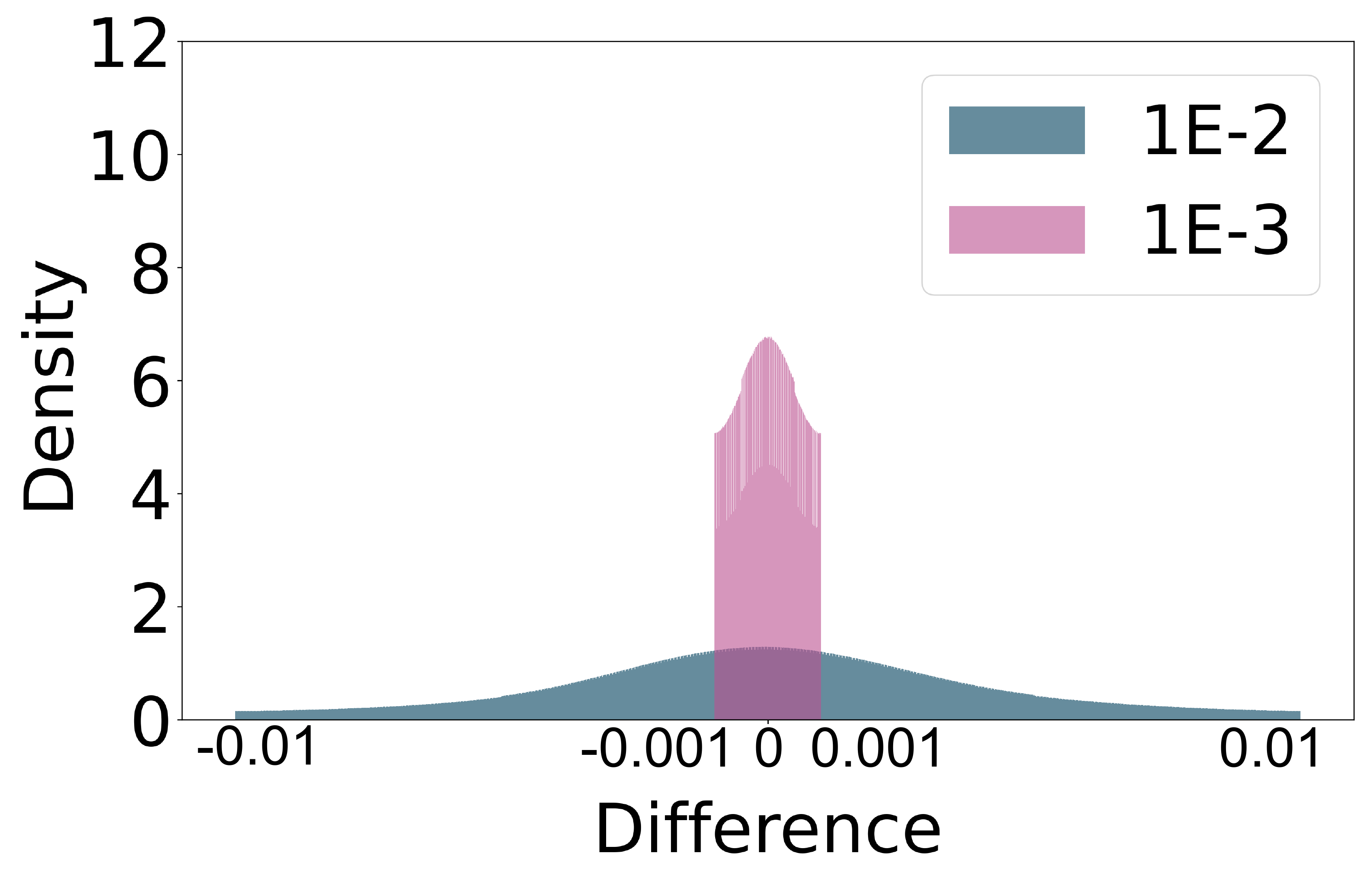}
        \caption{SZ 3.1}
    \end{subfigure}%
    \hfill
    \begin{subfigure}[b]{0.49\linewidth}
        \centering
        \includegraphics[width=\linewidth]{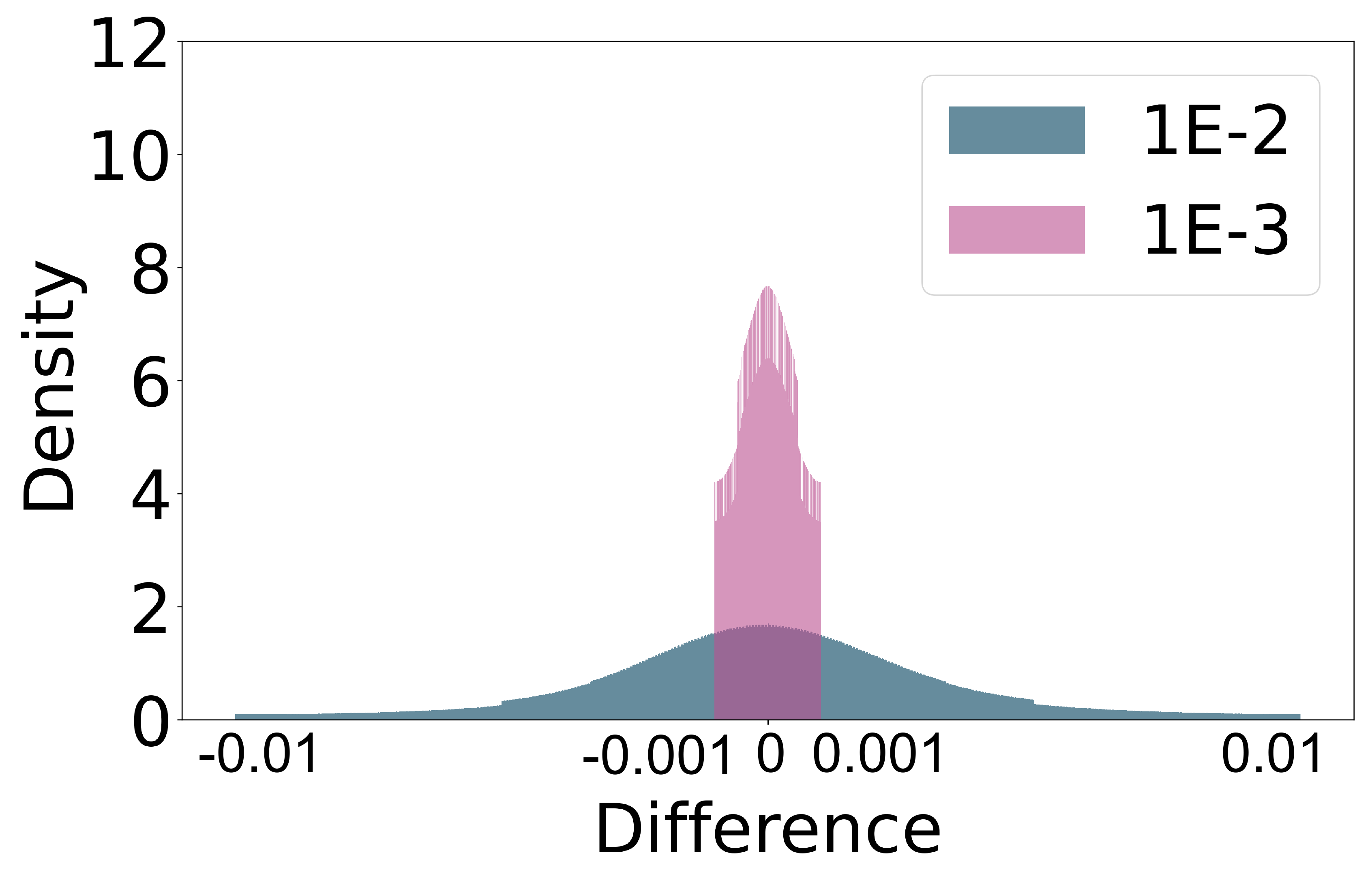}
        \caption{HPEZ}
    \end{subfigure}%
    \begin{subfigure}[b]{0.49\linewidth}
        \centering
        \includegraphics[width=\linewidth]{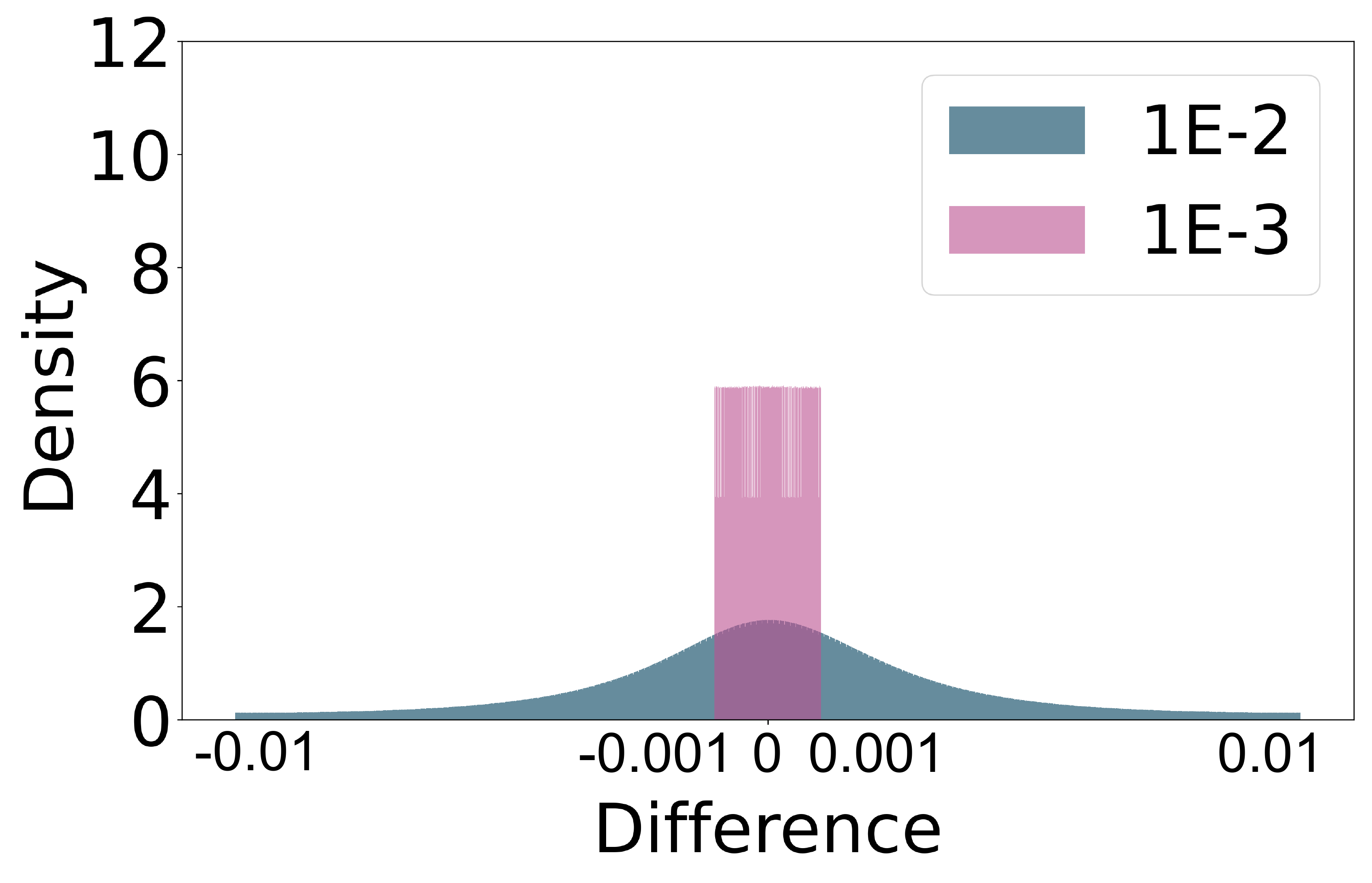}
        \caption{\GraphCom}\label{fig:exp:error-analysis-graphcom}
    \end{subfigure}%
    \caption{Decompression error distribution on \RS500.}
     \label{fig:exp:error-analysis}
\end{figure}

\stab
{\bf Error analysis.}
To understand the behaviour of each method, we analyse the decompression error\footnote{Due to the error-bounded module, all methods satisfy error bound $\epsilon$.}. Figure~\ref{fig:exp:error-analysis} shows the distribution of the point-wise relative decompression error for two bounds $\epsilon = 10^{-2}$ and $10^{-3}$.  Intuitively, we prefer few and small errors, resulting to a peak at the middle of the distribution. Observe that the distribution for {\GraphCom} is more spread out. Therefore, although {\GraphCom} achieves higher compression than its competitors, there is still room for improvement; we leave this as future work.

\begin{figure*}[tb]
    \centering
    \begin{subfigure}[b]{0.33\linewidth}
        \centering
        \includegraphics[width=\linewidth]{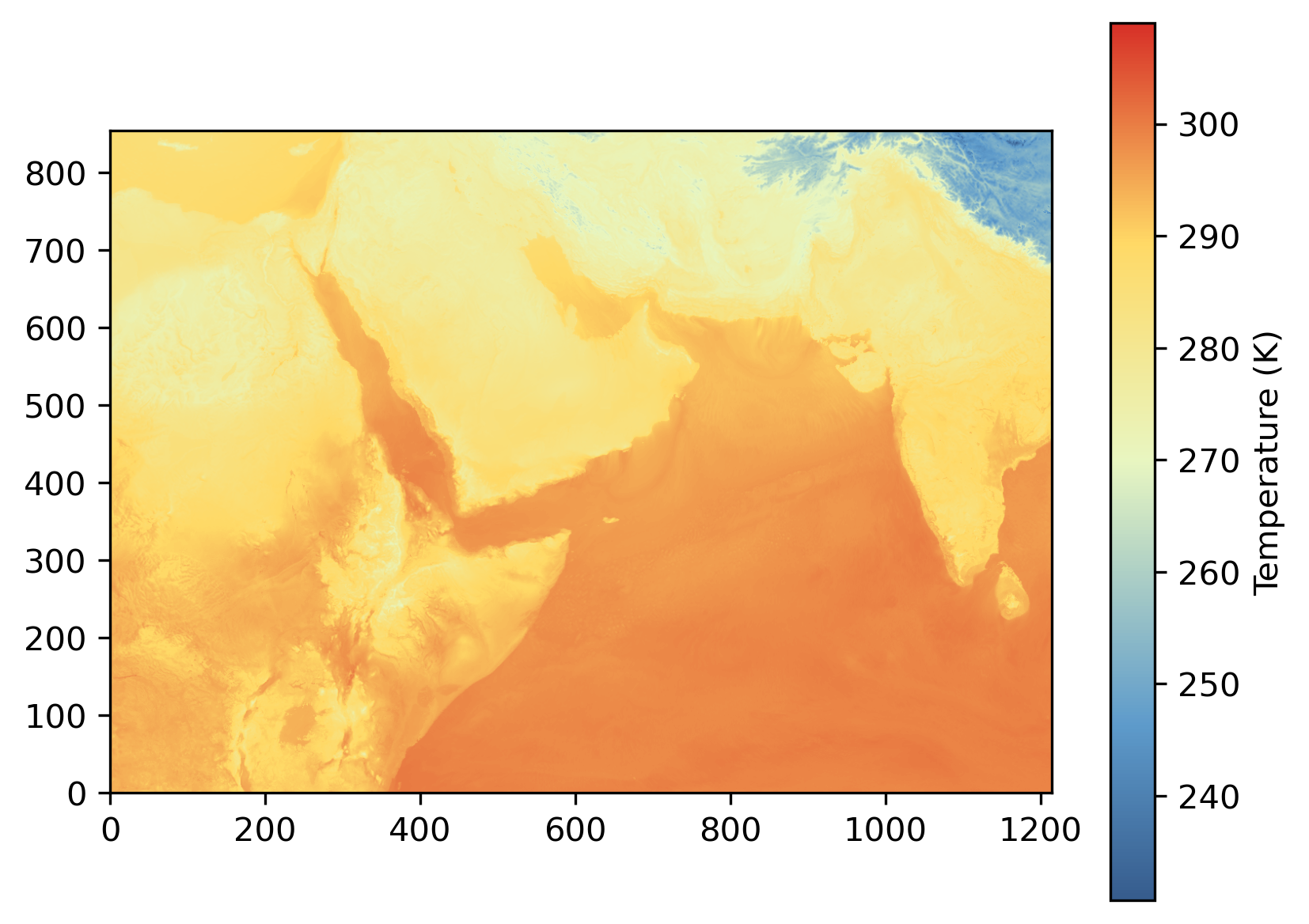}
        \caption{Original}\label{fig:exp:abs-diff-original}
    \end{subfigure}%
    \begin{subfigure}[b]{0.33\linewidth}
        \centering
        \includegraphics[width=.97\linewidth]{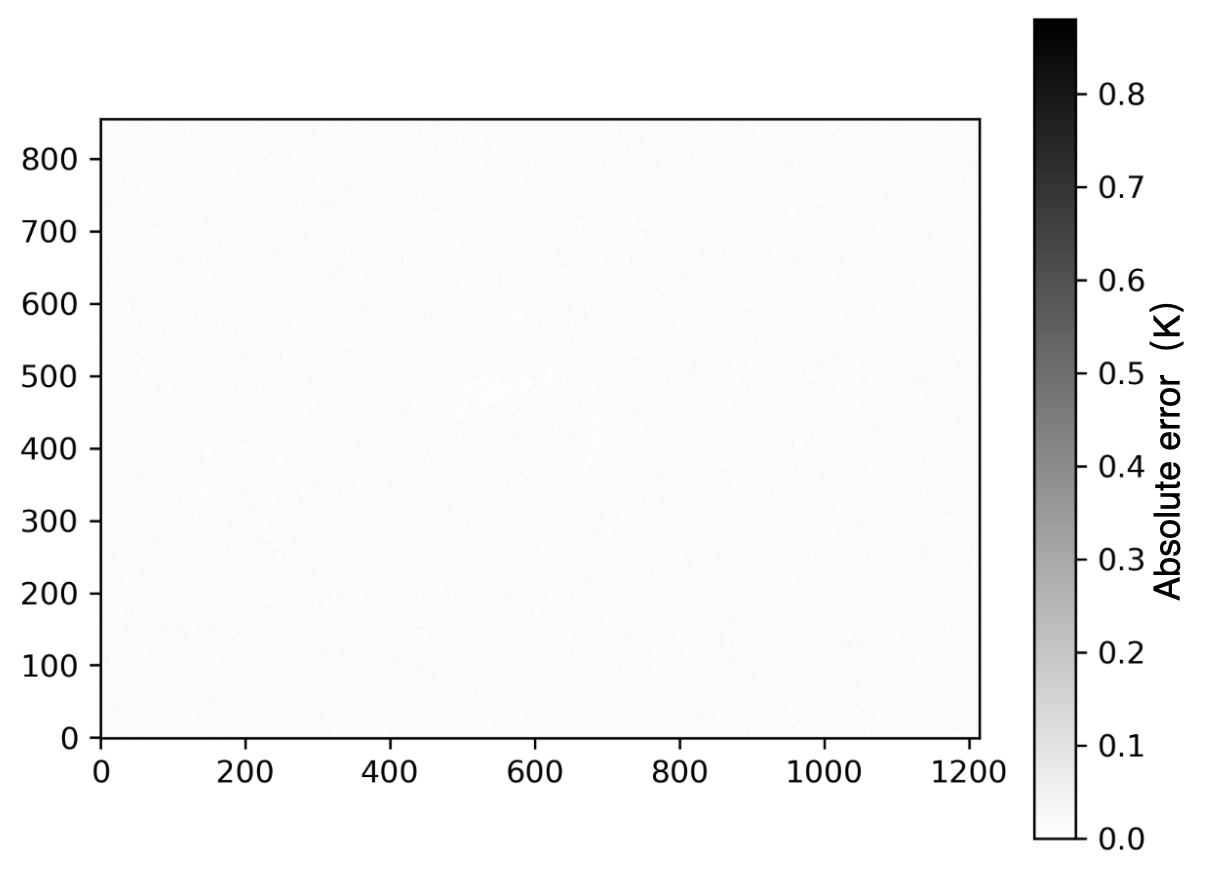}
        \caption{ZFP achieves $\rho = 8.53$}\label{fig:exp:abs-diff-zfp}
    \end{subfigure}%
    \begin{subfigure}[b]{0.33\linewidth}
        \centering
        \includegraphics[width=.97\linewidth]{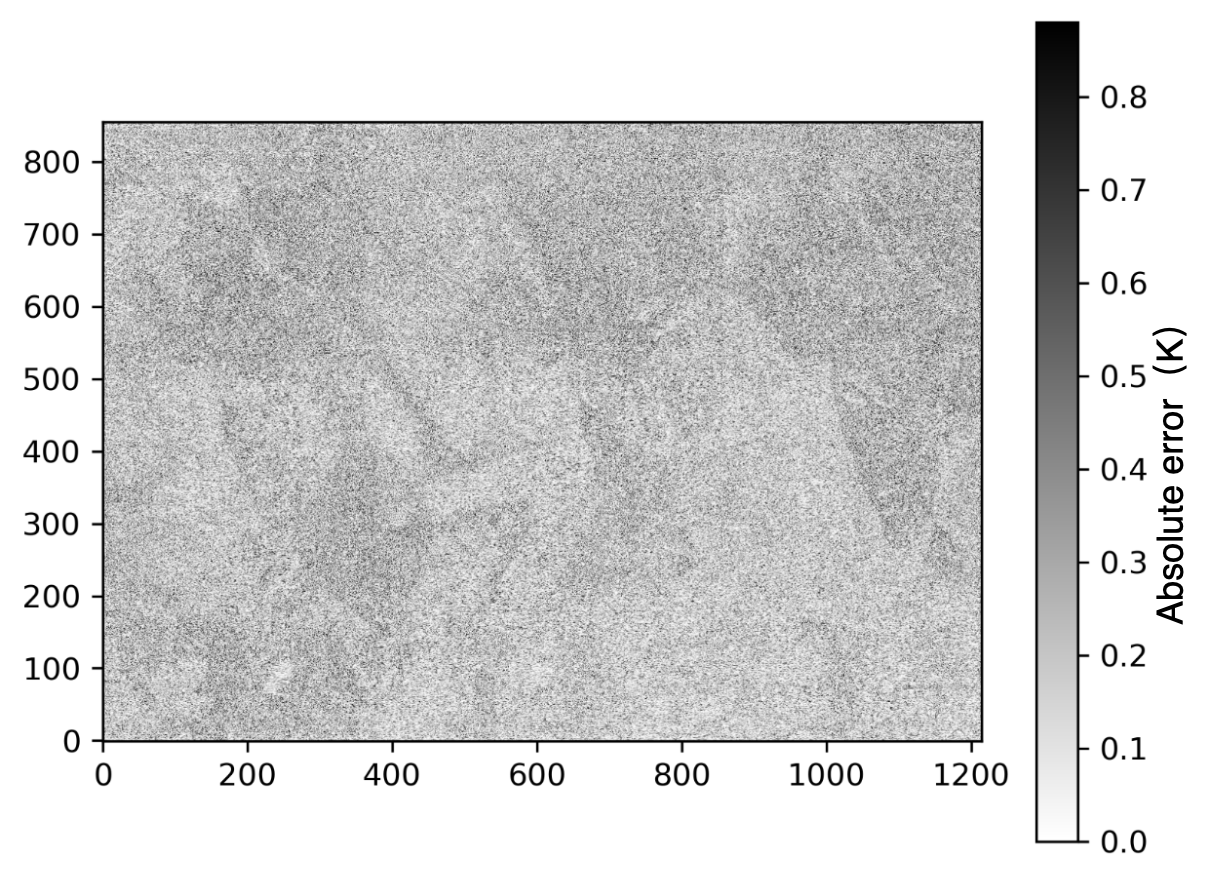}
        \caption{SPERR achieves $\rho = 92.54$}\label{fig:exp:abs-diff-sperr}
    \end{subfigure}%
    \hfill
    \begin{subfigure}[b]{0.33\linewidth}
        \centering
        \includegraphics[width=.97\linewidth]{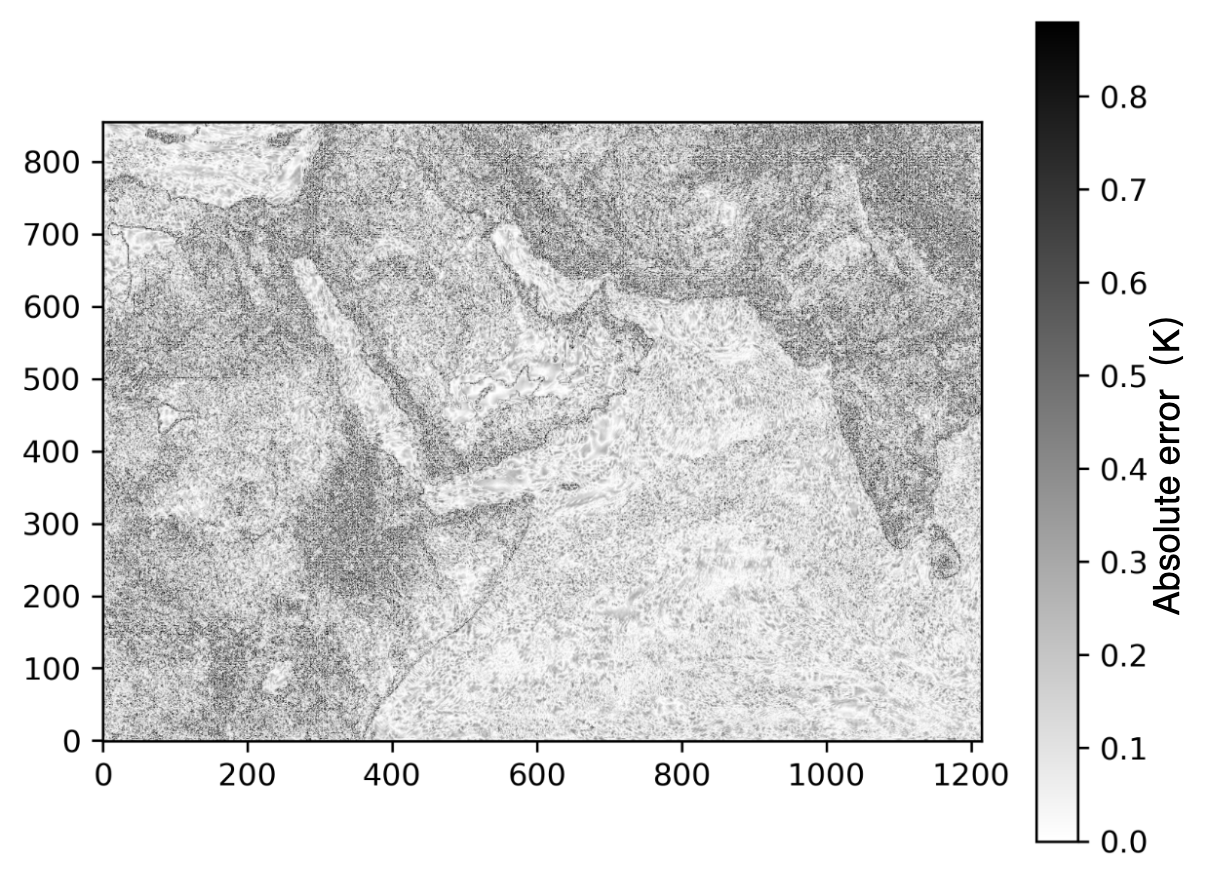}
        \caption{HPEZ achieves $\rho = 102.59$}\label{fig:exp:abs-diff-hpez}
    \end{subfigure}%
    \begin{subfigure}[b]{0.33\linewidth}
        \centering
        \includegraphics[width=.97\linewidth]{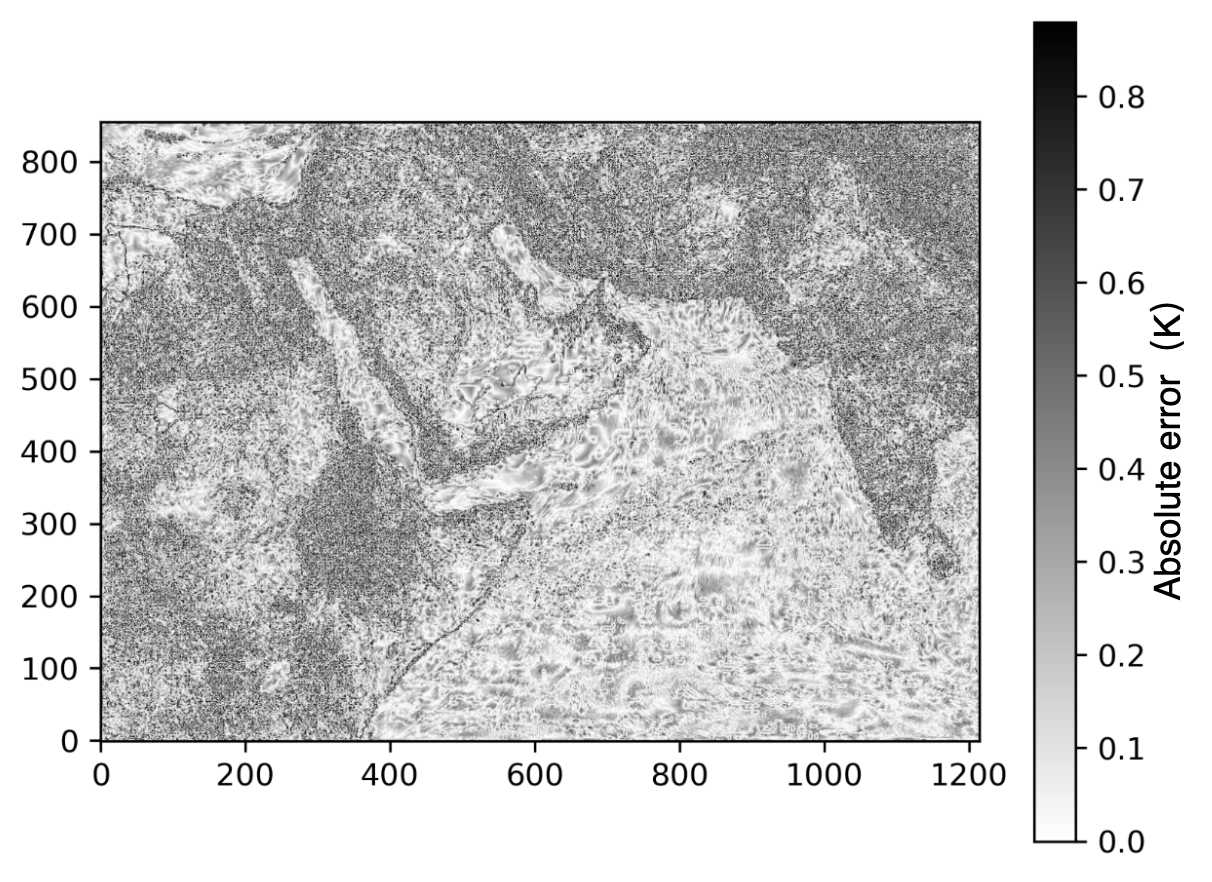}
        \caption{SZ 3.1 achieves $\rho = 105.21$}\label{fig:exp:abs-diff-sz}
    \end{subfigure}%
    \begin{subfigure}[b]{0.33\linewidth}
        \centering
        \includegraphics[width=.97\linewidth]{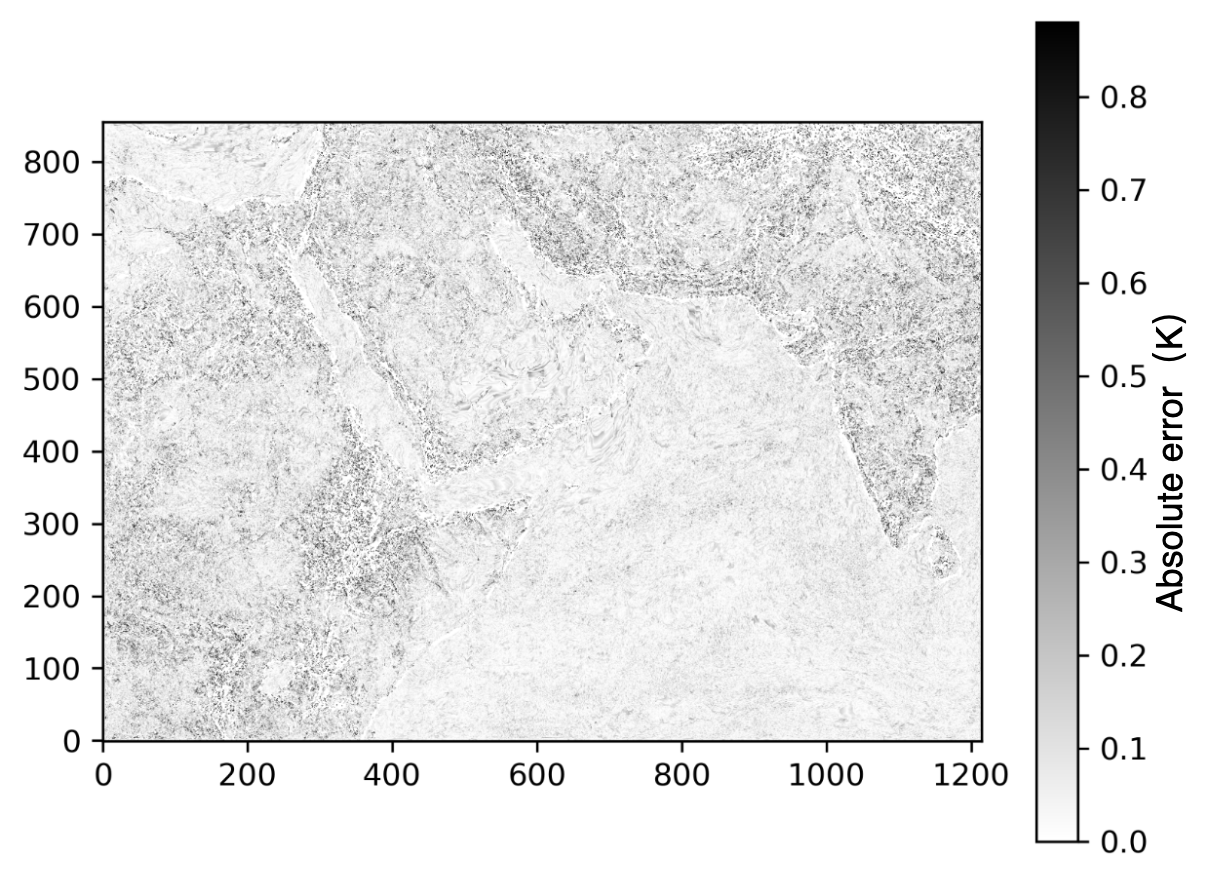}
        \caption{{\GraphCom} achieves $\rho = 141.75$}\label{fig:exp:abs-diff-graphcom}
    \end{subfigure}%
    \caption{Absolute difference between original and decompressed data for the \RS96K datast ($\epsilon = 10^{-2}$); darker color represents larger error}
    \label{fig:exp:reconstructed-temperature-data-t0-on-different-methods}
\end{figure*}

Figure~\ref{fig:exp:reconstructed-temperature-data-t0-on-different-methods} visualizes the absolute point-wise error for a random timestamp of the \RS96K (\ie temperature) dataset, for $\epsilon = 10^{-2}$; darker color represents larger error. ZFP is a fixed-rate technique generating few very small decompression errors; the trade-off is its low compression ratio $\rho = 8.53$. On the other hand, the error of SPERR is larger, but $\rho = 92.54$ is also higher. Observe that the distribution of error in SPERR is uniform, in contrast to SZ3.1 whose error is biased towards areas with land masses. \ie Eastern Africa, the Arabic Peninsula and India are clearly highlighted. Finally, {\GraphCom} results in moderate error that is rather uniformly distributed, and achieves the highest $\rho = 141.75$.      

\begin{figure}[t]
    \centering
    \begin{subfigure}[b]{0.8\linewidth}
        \centering
        \includegraphics[width=\linewidth]{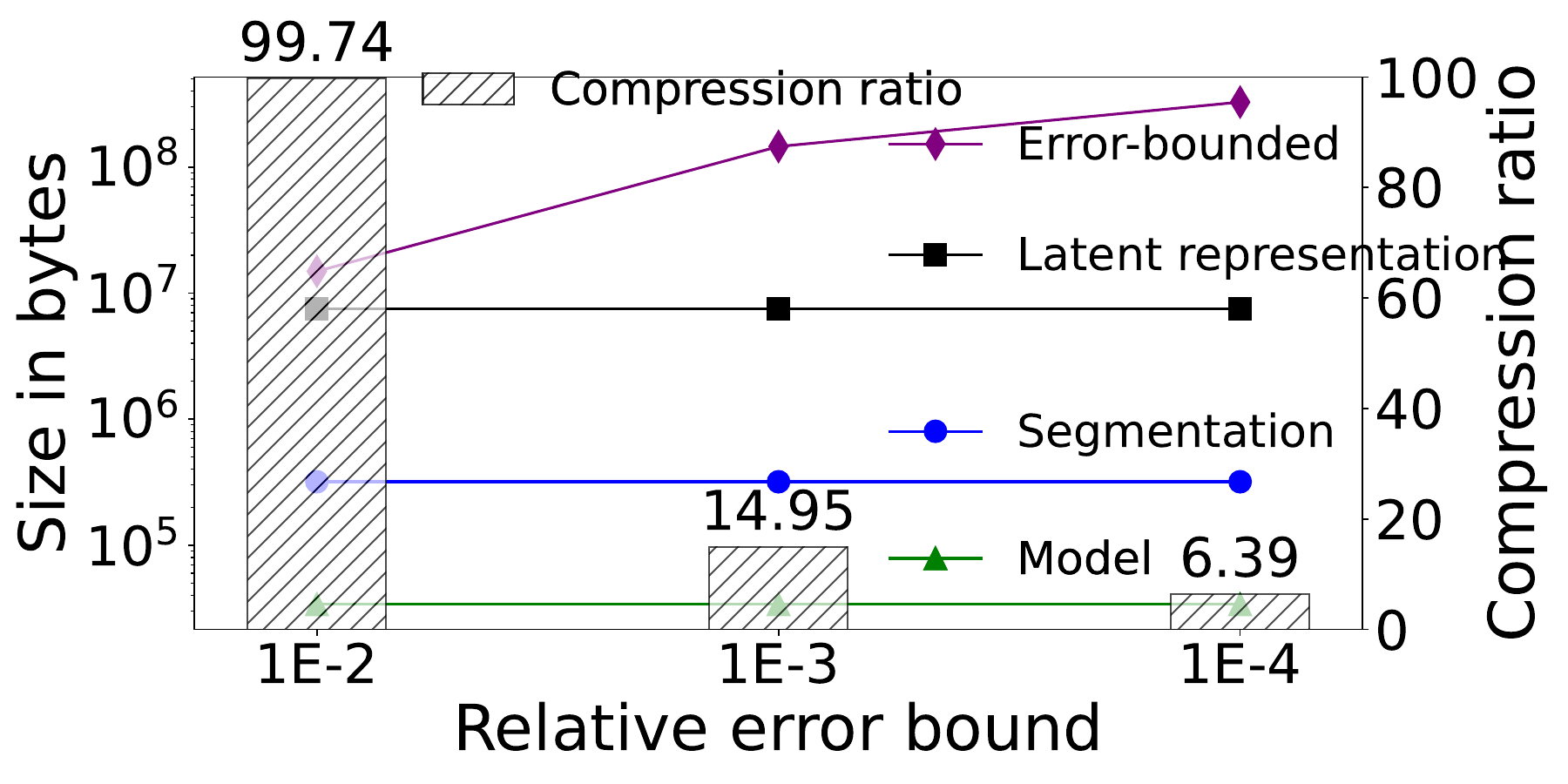}
    \end{subfigure}%
    \caption{Size in bytes of the segmentation, model, latent representation and residual error component of {\GraphCom} for the \RS500 dataset ($y$-axis on log-scale). Bars represent the compression ratio $\rho$.}
    \label{fig:exp:cr-four-components}
\end{figure}

\subsection{Ablation Study}\label{exp:cr-four-components}

\noindent
{\bf Size of compressed data components.}
{\GraphCom} stores the compressed data as four separate components: \myNum{i} segmentation, that stores the spatial extents of the regions generated by the the Felzenszwalb algorithm for virtual timestamp $X_\mu$; 
\myNum{ii}~model weights for autoencoder {\GNN}; 
\myNum{iii} latent representation of the graphs in set $\mathcal{G}$; and
\myNum{iv} compressed residual error information, which is used by the error-bounded module to correct errors. 

Figure~\ref{fig:exp:cr-four-components} shows the size in bytes for each component, for the \RS500 dataset and three error bounds $\epsilon = 10^{-2}$, $10^{-3}$ and $10^{-4}$. Note that the $y$-axis is in log-scale. 
The segmentation, model and latent representation components are relatively small, \ie less than 10MB, where the size of the original data is 1.9GB. Moreover, the size of these components does not vary with $\epsilon$.  
By far, the largest component is the residual error information, which also increases for tighter bounds $\epsilon$. 

In Figure \ref{fig:exp:cr-four-components} we use bars to display the compression ratio $\rho$. There is correlation between the size of the residual error information and $\rho$. To investigate further, 
in the following table we report the percentage of points generated by {\GraphCom} during grid reconstruction, that do not satisfy $\epsilon$. 

\begin{center}
\small
\begin{tabular}{lccc}
\hline
  \bf Relative error bound $\epsilon$ & $10^{-2}$ & $10^{-3}$  & $10^{-4}$   \\\hline\hline
  \bf Ratio of erroneous predictions  & 2.72\%   & 34.62\%  &  47.71\% \\
  \bf Compression ratio $\rho$    &  99.74    & 14.95     & 6.39   \\
\hline
\end{tabular}
\end{center}

Those points must be subsequently corrected; this increases the size of the residual error information, thus decreasing $\rho$. For example, when $\epsilon = 10^{-2}$, only 2.72\% of the reconstructed points need correction.  

\begin{figure}[t]
    \centering
    \begin{subfigure}[b]{0.67\linewidth}
        \centering
        \includegraphics[width=\linewidth]{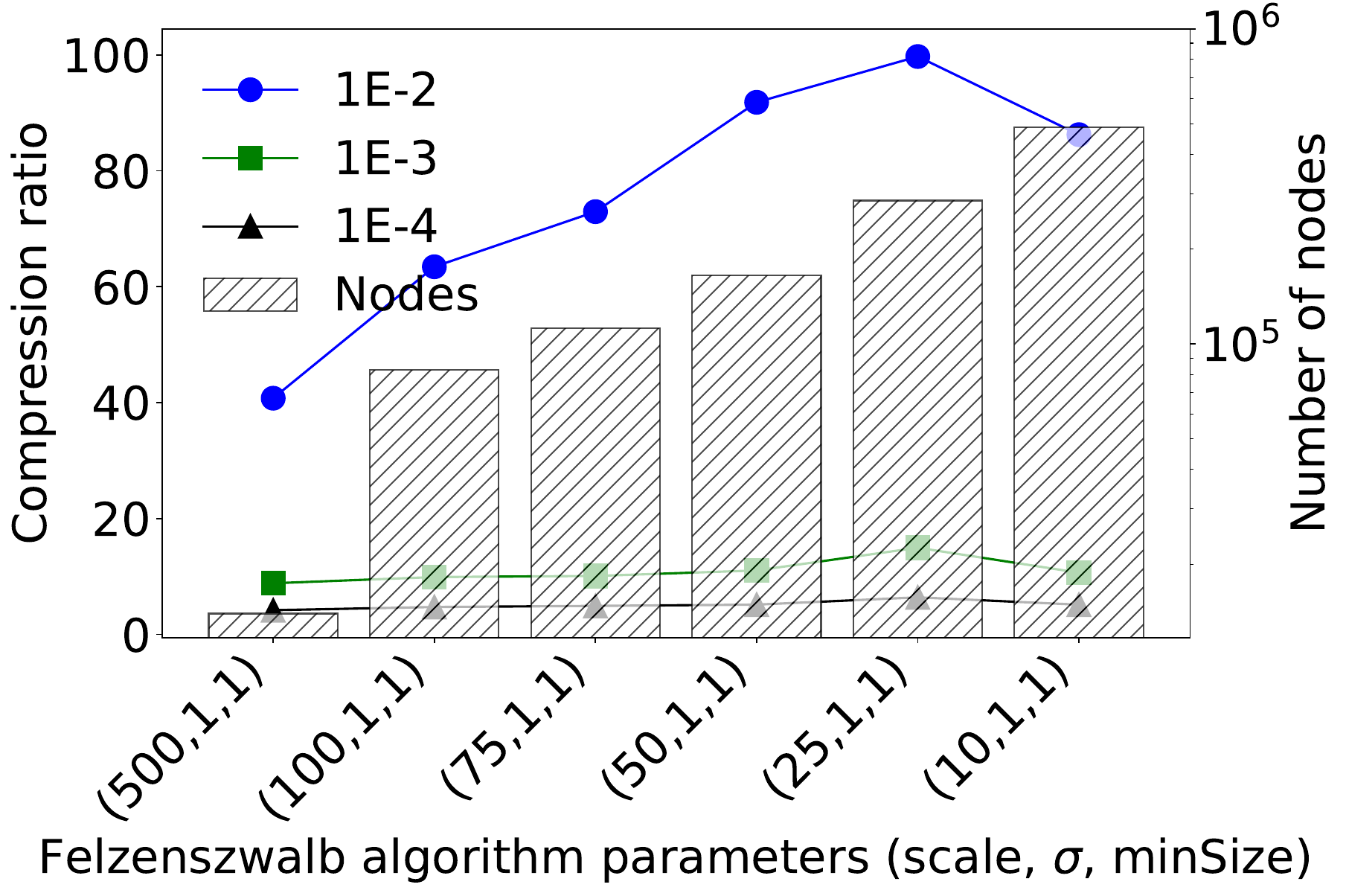}
    \end{subfigure}%
    \caption{Compression ratio $\rho$ of the \RS500 dataset for various combinations of parameters $\langle \text{scale}, \sigma, \text{minSize} \rangle$ for the Felzenszwalb segmentation algorithm. Each line corresponds to a different relative error bound $\epsilon$. Bars represent the number of nodes.}
    \label{fig:exp:cr-parameters-segmentation}
\end{figure}

\stab
{\bf Segmentation parameters.}
We evaluate the impact of parameters 
$\langle \text{scale}, \sigma, \text{minSize} \rangle$
of the Felzenszwalb algorithm~\cite{felzenszwalb2004efficient} used for segmentation on the compression ratio, using the \RS500 dataset. 
These parameters determine the number of regions generated during segmentation, which in turn correspond to the number of nodes in every graph $G_t \in \mathcal{G}$.

Figure~\ref{fig:exp:cr-parameters-segmentation} presents the compression ratios achieved with various combinations of these parameters.
The settings range from $\langle 500, 1, 1\rangle$ to $\langle 1, 1, 1\rangle$, leading to significant variations in the number of graph nodes, from tens of thousands to hundreds of thousands; consequently affecting the segmentation granularity.
The results show that as the number of nodes (depicted by the bar chart) increases, the compression ratio also increases, peaking at the $\langle 10, 1, 1\rangle$ configuration, before decreasing. 
Intuitively, larger number of nodes corresponds to more detailed segmentation; therefore the subsequent grid reconstruction process is more accurate and the resulting residual error information is smaller. However, after a certain number of nodes, the improvements in terms of accurate reconstruction are diminishing, whereas the size of the latent representation grows disproportionately, reducing the compression ratio.    
The $\langle 10, 1, 1\rangle$ combination consistently yields the highest compression ratio, therefore it is used as default for all other experiments.

\begin{figure}[t]
    \centering
    \begin{subfigure}[b]{0.6\linewidth}
        \centering
        \includegraphics[width=\linewidth]{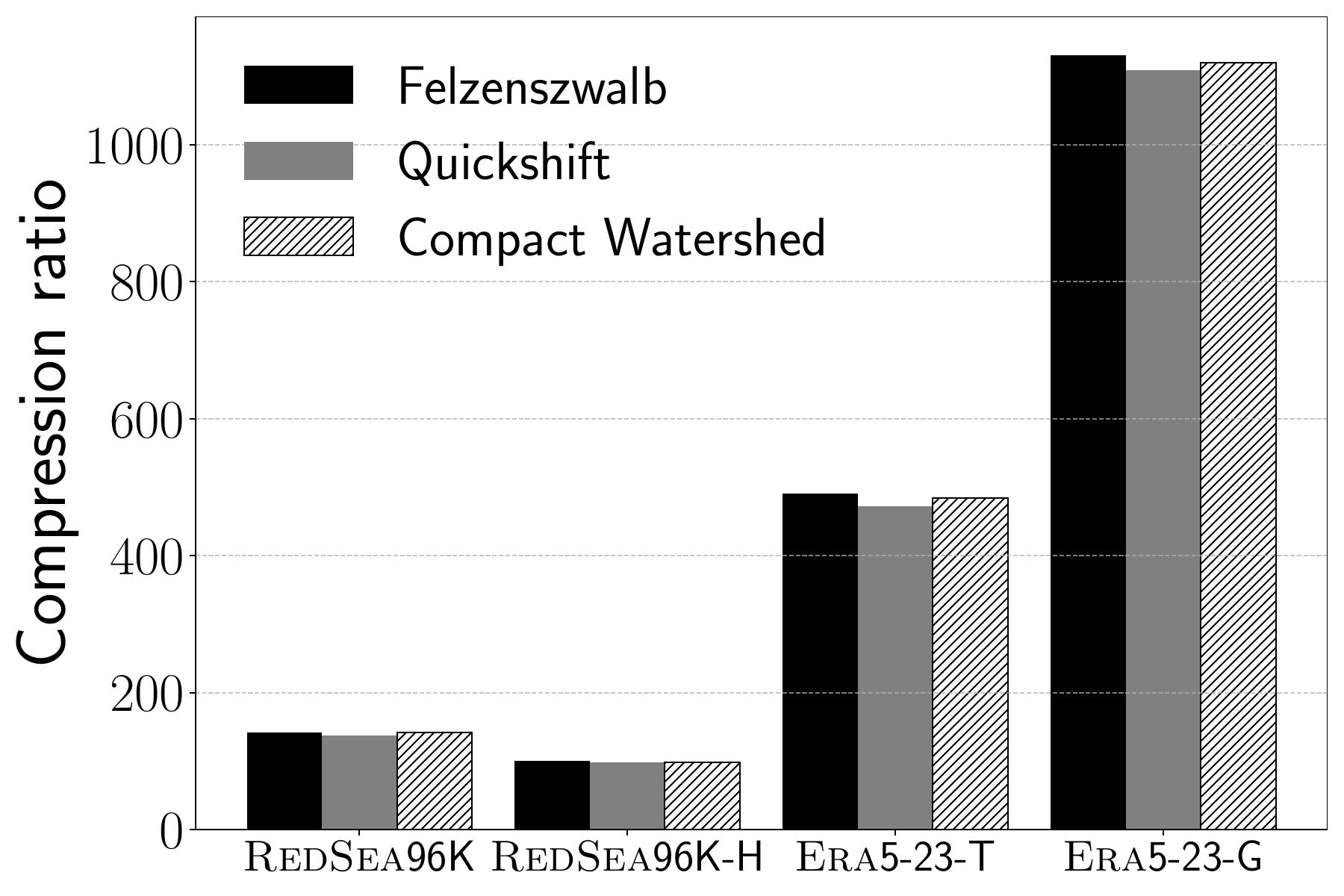}
    \end{subfigure}%
    \caption{Compression ratio for different segmentation algorithms ($\epsilon = 10^{-2}$).}
    \label{fig:cr-different-segmentation-algorithm}
\end{figure}

\stab
{\bf Segmentation algorithms evaluation.}\label{exp:different-segmentation-algos}
We evaluate the effect of different segmentation algorithms on the compression ratio $\rho$. We use three algorithms, namely Felzenszwalb~\cite{felzenszwalb2004efficient} (our default choice), Quickshift~\cite{vedaldi2008quick}, and Compact Watershed~\cite{neubert2014compact}.
Figure~\ref{fig:cr-different-segmentation-algorithm} presents the compression ratio achieved by each segmentation algorithm.
The results indicate that compression ratios are generally consistent across all segmentation algorithms for each dataset.
This consistency arises from the similar number of nodes produced by the different algorithms, resulting in a limited impact on the overall compression ratio.

\begin{figure}[t]
    \centering
    \begin{subfigure}[b]{0.65\linewidth}
        \centering
        \includegraphics[width=\linewidth]{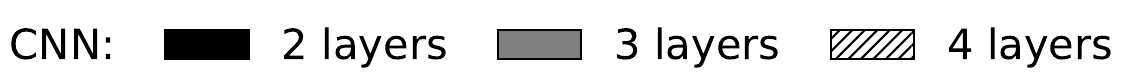}
    \end{subfigure}
    \begin{subfigure}[b]{0.49\linewidth}
        \centering
        \includegraphics[width=\linewidth]{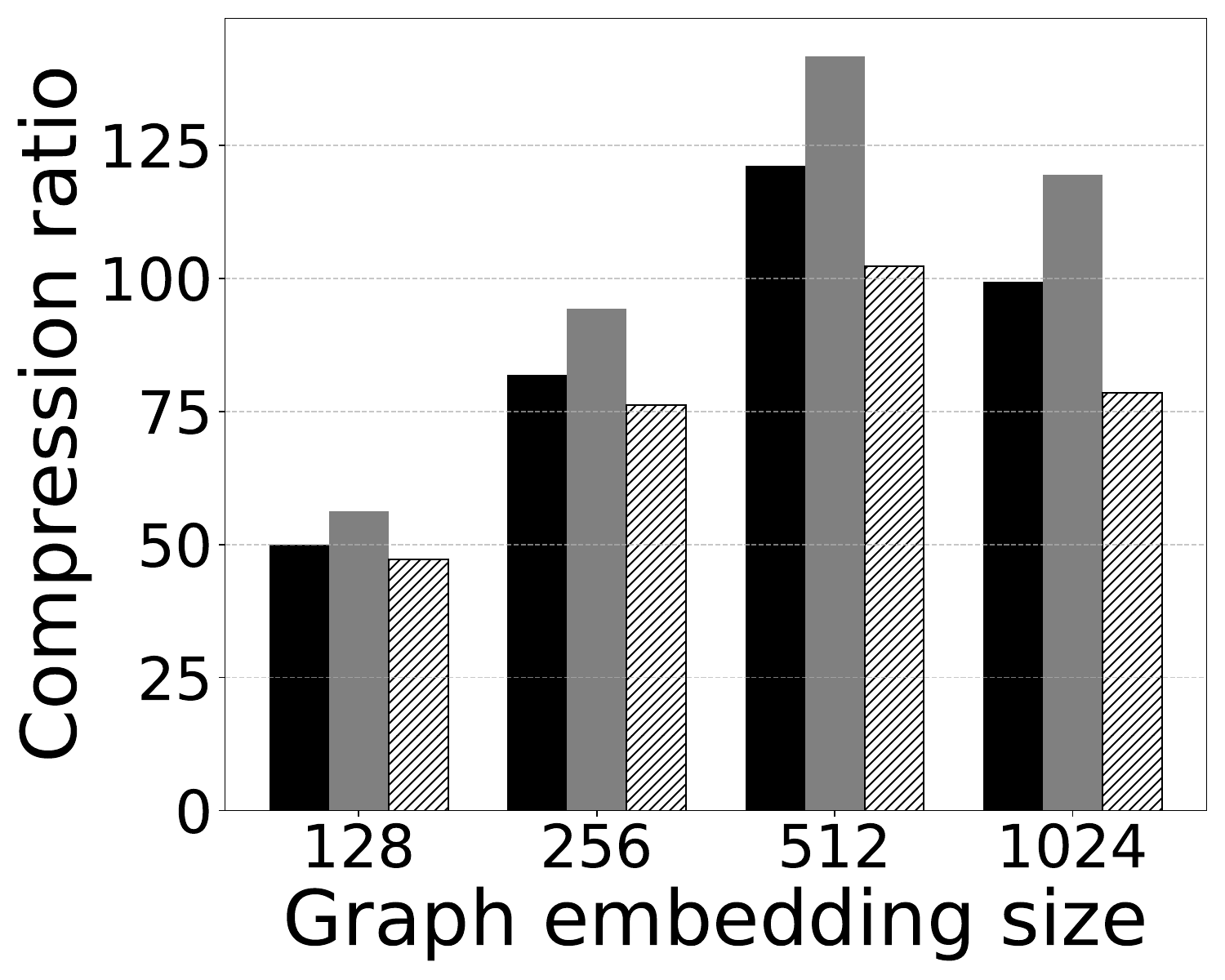}
        \caption{\RS96K}
        \label{fig:cr-latent-redsea96k}
    \end{subfigure}
    \begin{subfigure}[b]{0.49\linewidth}
        \centering
        \includegraphics[width=\linewidth]{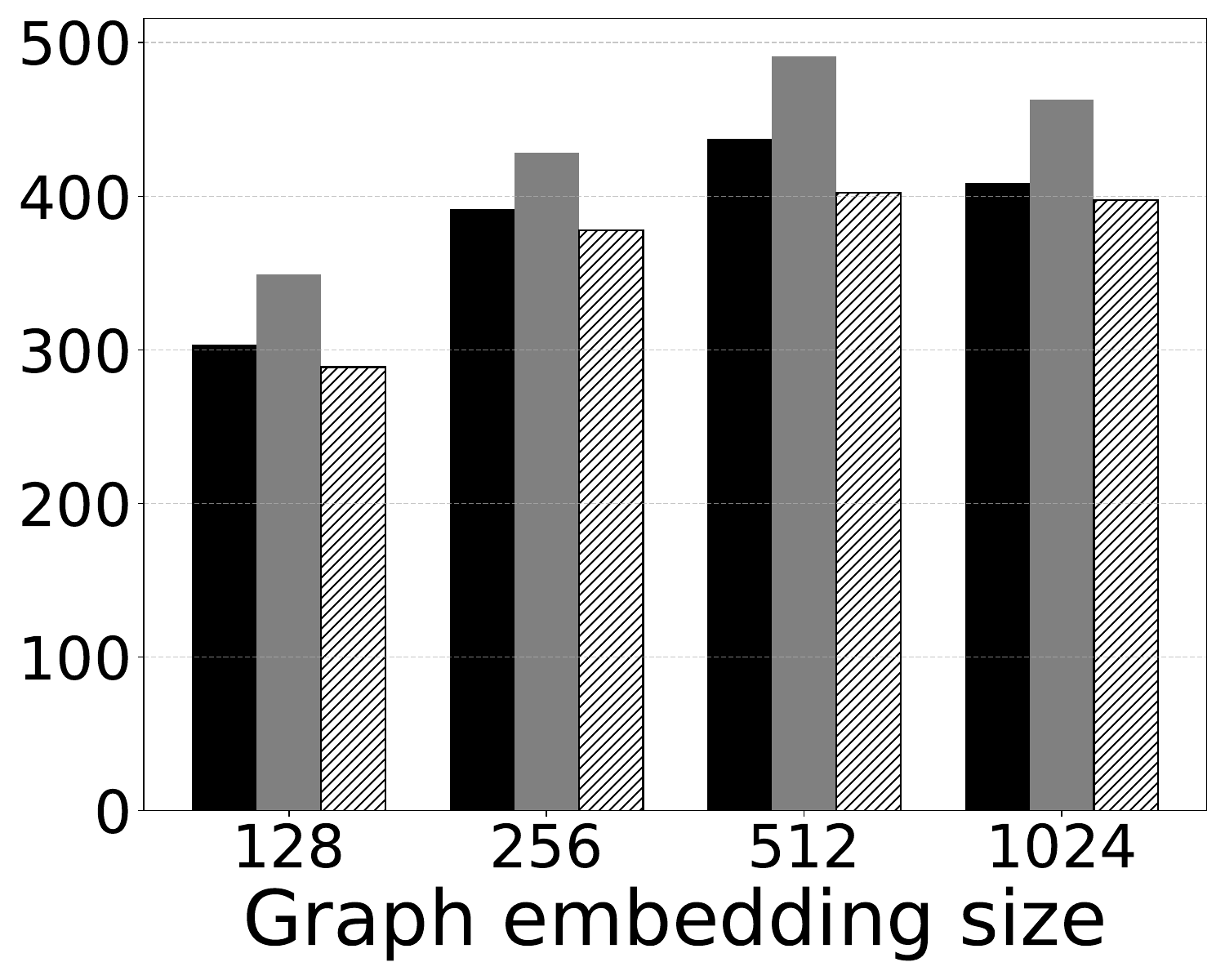}
        \caption{\ERA5-23-T}
        \label{fig:cr-latent-era5-2023T}
    \end{subfigure}
    \caption{Compression ratios across different graph embedding size and CNN layers ($\epsilon = 10^{-2}$).}
    \label{fig:cr-different-latent-size}
\end{figure}

\stab
{\bf Varying spatial and temporal features.}
We now investigate how spatial (graph embedding size) and temporal (CNN layers) features influence the compression ratio $\rho$. Both these factors critically affect compression ratio.
Specifically, graph embedding size determines spatial granularity, while CNN layers capture temporal dependencies.
As shown in Figure~\ref{fig:cr-different-latent-size}, the compression ratio $\rho$ reaches its peak at a graph embedding size of 512 with 3 CNN layers, achieving a balance between spatial representation and temporal feature extraction.
Notably, $\rho$ decreases significantly with smaller embeddings due to their limited spatial capacity, leading to higher residual errors and a reduced overall compression ratio.

\subsection{WGAN: Synthetic temperature data}\label{sec:method:wgan-redsea}

In this section, we introduce a generative adversarial network, specifically a Wasserstein GAN (WGAN)~\cite{arjovsky2017wasserstein}, to generate temperature data that reflects the characteristics of our proprietary {\RS} dataset \cite{hoteit-RSRA2018, hoteit-RSRA2022}.

Although there exist many similar public real datasets (like the \ERA5~\cite{hersbach2020era5} family), they typically lack two characteristics: \myNum{i} the number of timestamps per region is limited; and \myNum{ii} their spatial resolution is coarse. Contrary {\RS} consists of more than 96K timestamps and represents the data at a fine spatial resolution. Consequently, compression is more challenging and stresses the limits of the tested methods. For example, recall from Table~\ref{table:final-cr-real-gan} that the highest compression ratio achieved by any method for {\RS} is $\rho \simeq 134$, whereas for \ERA5 the best ratio reaches $\rho \simeq 1,073$.

Given the proprietary nature of our {\RS} data, direct sharing is infeasible. However, by using our WGAN-based generator, we can provide synthesized data that maintains the statistical characteristics of the original dataset. Moreover, the WGAN model can generate an expanded volume of synthetic data, which is vital for testing the efficacy of our compression method on more extensive datasets. For these reasons, we believe our WGAN generator is an important contribution both for reproducibility and as a service to the academic community.

To ensure the utility of WGAN-generated data in compression studies, we adopt a rigorous training process. This includes the implementation of a gradient penalty method to stabilize training and optimizing both the generator and critic~\cite{arjovsky2017wasserstein} components. We then fine-tune the WGAN model to ensure that the generated data closely resembles the real data. We configure the batch size to $32$, the number of epoch to $10^4$, and the learning rate to $8\cdot 10^{-5}$. When generating synthetic data, the number of timestamps can be adjusted to accommodate the full dataset, or to generate a smaller subset that fits within the GPU memory constraints. 

To validate the similarities and differences between real and WGAN-generated data, we employ a comprehensive set of evaluation strategies, including visual comparisons, statistical properties comparison to quantify the correspondence between datasets, and benchmark performance testing using all aforementioned compression methods. The evaluation methods collectively provide a robust assessment of the relationship between the real and WGAN-generated datasets.

\begin{figure}
  \centering
  \begin{minipage}[b]{0.5\textwidth}
    \centering
    \includegraphics[width=0.49\linewidth]{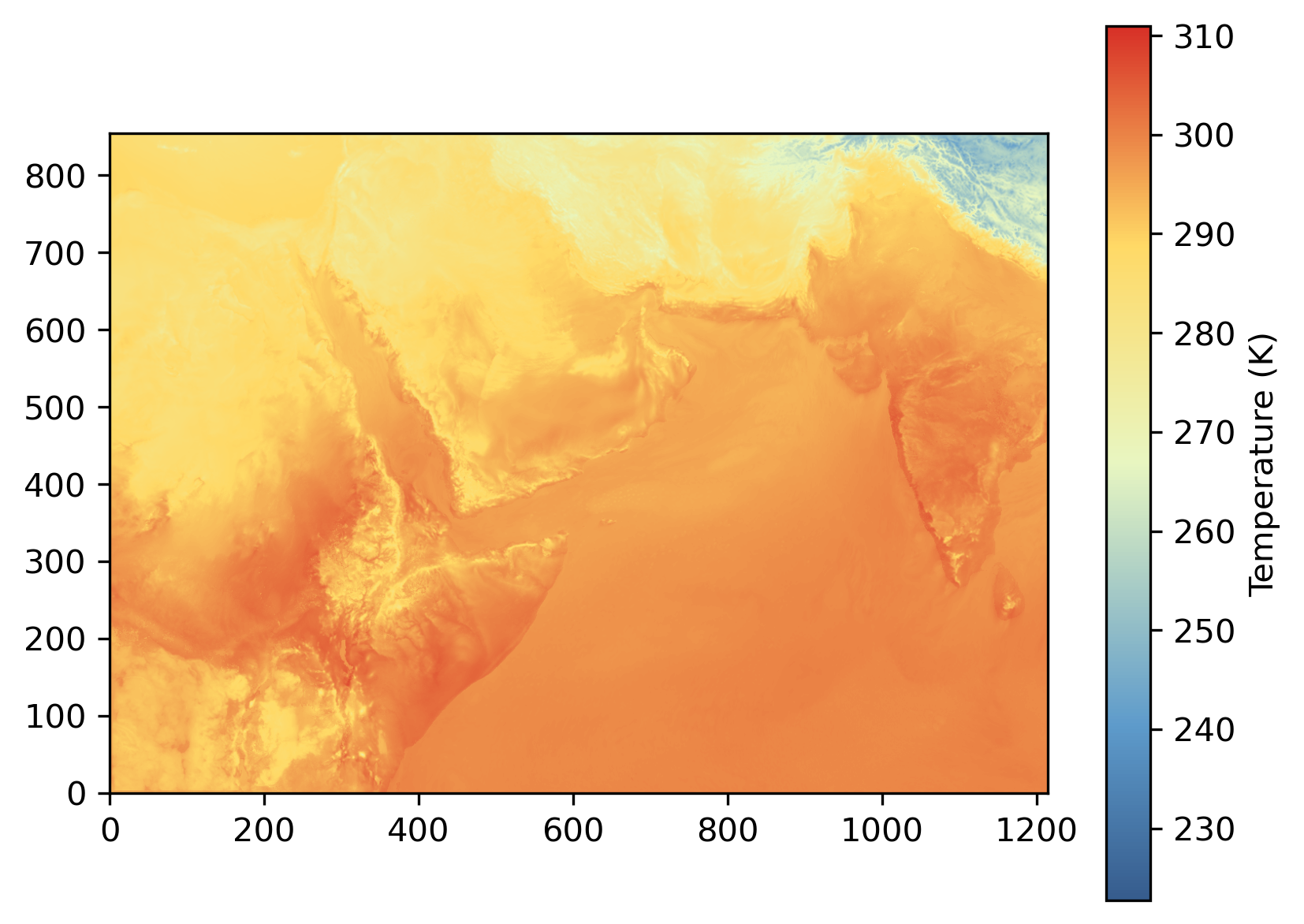}
    \includegraphics[width=0.49\linewidth]{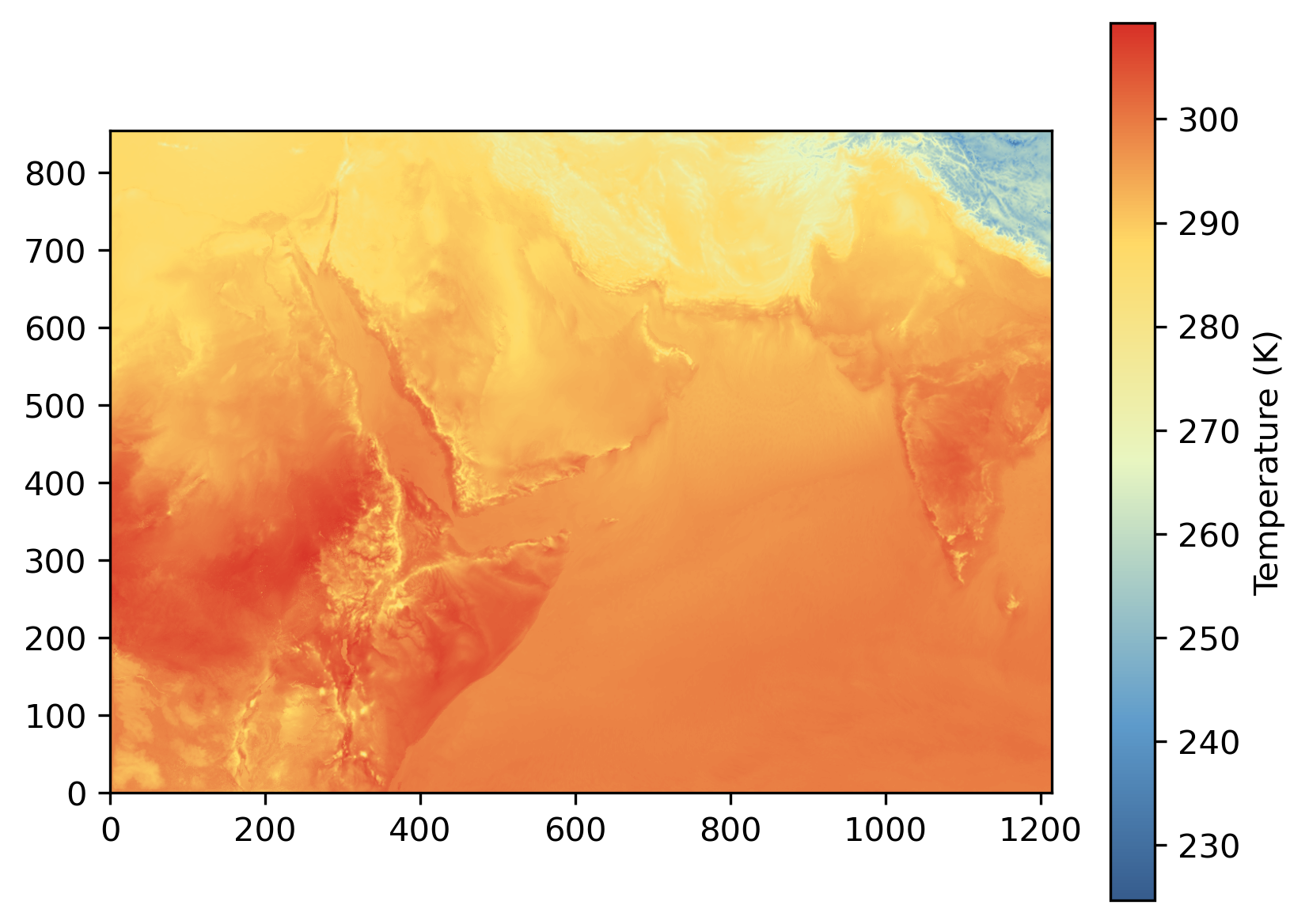}
    \subcaption{Real data}
  \end{minipage}%
  \hfill
  \centering
  \begin{minipage}[b]{0.5\textwidth}
    \centering
    \includegraphics[width=0.49\linewidth]{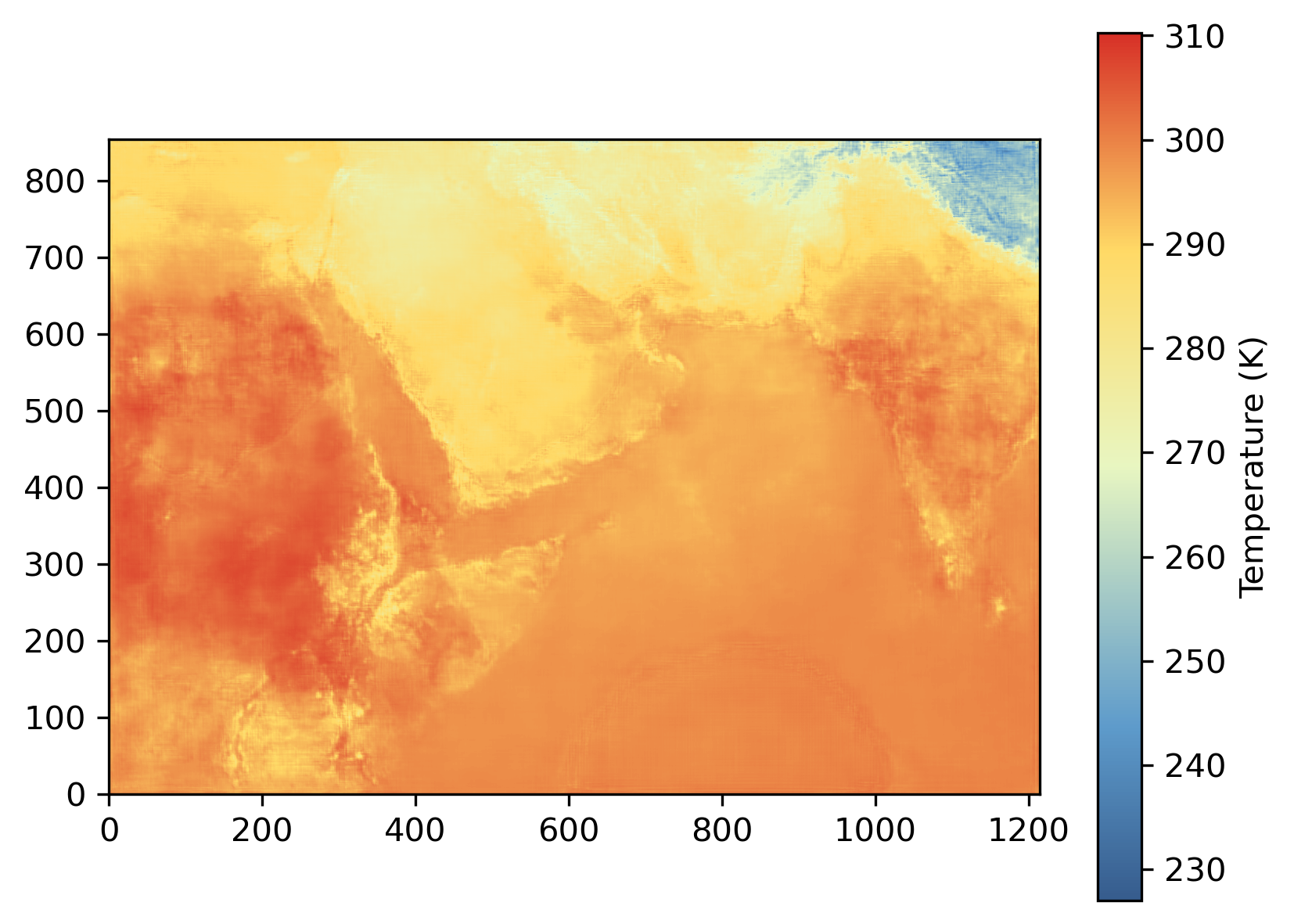}
    \includegraphics[width=0.49\linewidth]{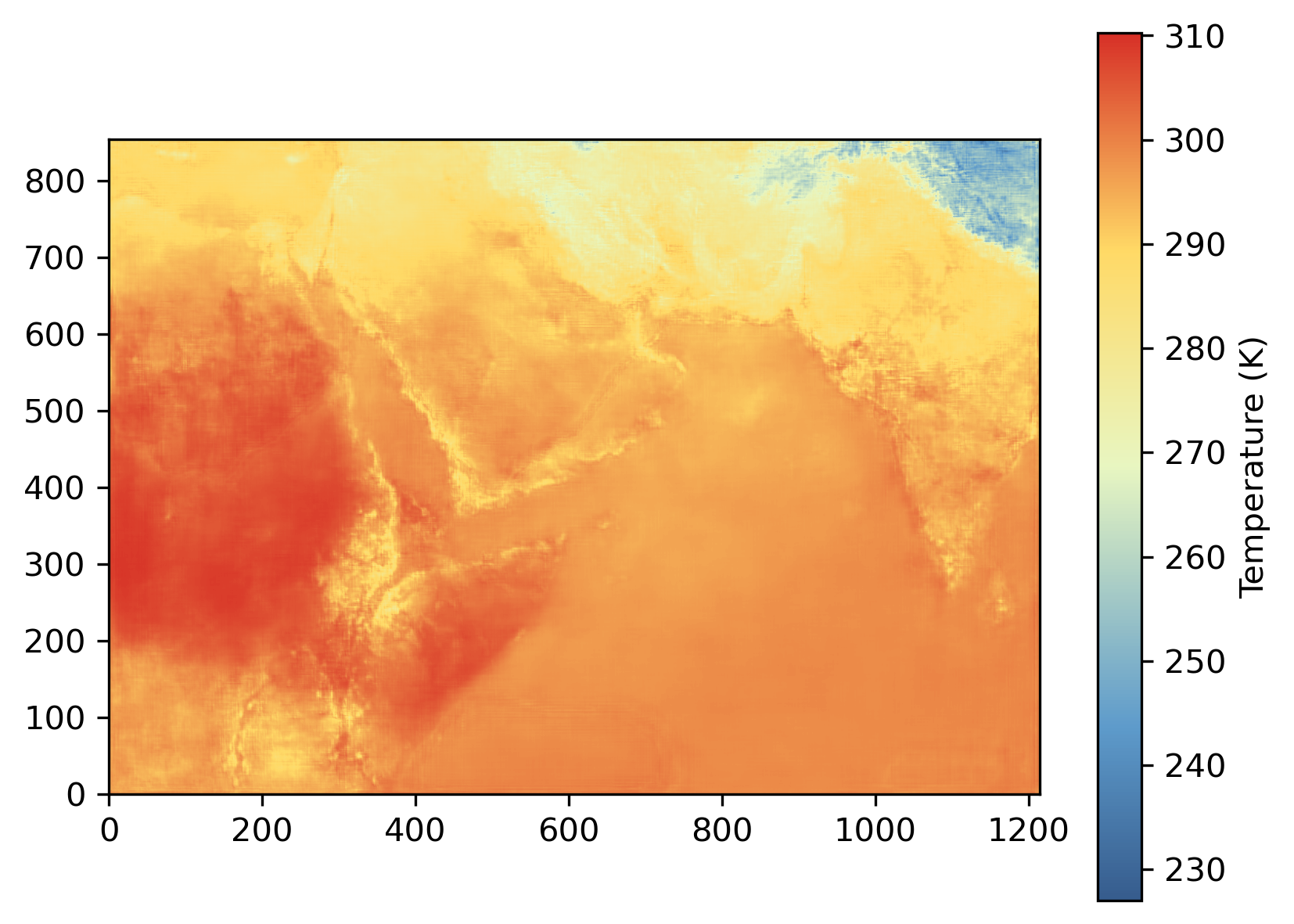}
    \subcaption{WGAN-generated data}\label{fig:GAN-generated_data}
  \end{minipage}%
  \hfill
  \centering
  \begin{minipage}[b]{0.5\textwidth}
    \centering
    \includegraphics[width=0.49\linewidth]{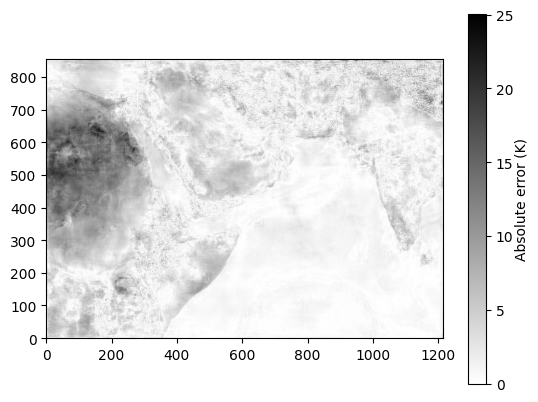}
    \includegraphics[width=0.49\linewidth]{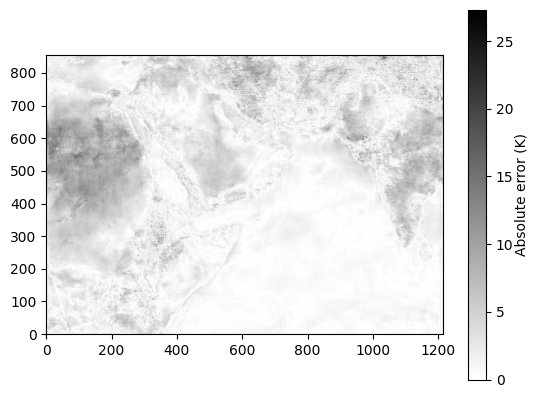}
    \subcaption{Absolute difference between real and WGAN-generated data}
    \label{fig:abs-difference-real-GAN}
  \end{minipage}%
  \caption{Comparative visualization of the real and WGAN-generated data.}\label{fig:comparison_real_and_generator}
\end{figure}

\stab
\noindent{\bf Visual comparison.}
We randomly select two timestamps from both the real and WGAN-generated datasets for analysis. Figure~\ref{fig:comparison_real_and_generator} presents a comparison between the real and WGAN-generated temperature data. 
The images show that our generator has successfully replicated the temperature data of specific regions, including the Arabian Peninsula, Red Sea, East Africa, India, and the Indian Ocean, as depicted in Figure~\ref{fig:GAN-generated_data}.
\guozhong{The absolute difference between real and WGAN-generated data is shown in Figure~\ref{fig:abs-difference-real-GAN} where darker color corresponds to larger absolute error.}

\begin{figure*}[t]
  \centering
  \begin{subfigure}[b]{0.33\textwidth}
    \centering
    \includegraphics[width=\linewidth]{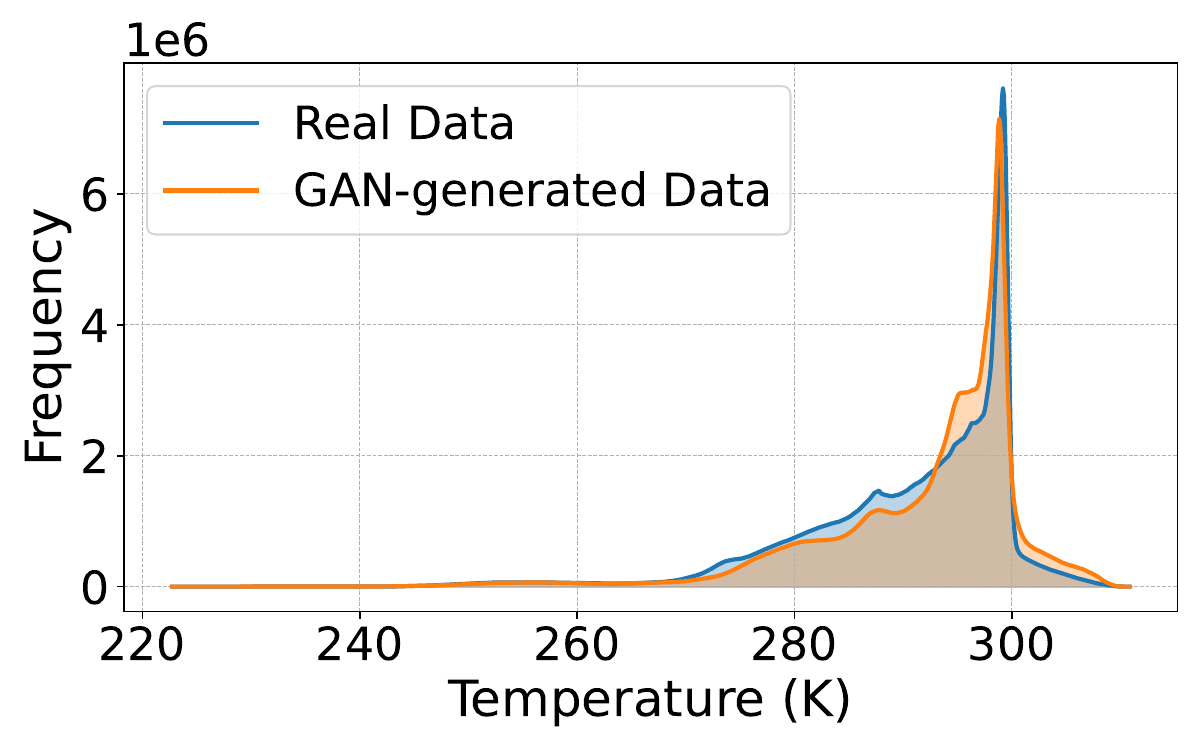}
    \caption{500 timestamps}
    \label{fig:distribution_500}
  \end{subfigure}%
  \hfill
  \begin{subfigure}[b]{0.33\textwidth}
    \centering
    \includegraphics[width=\linewidth]{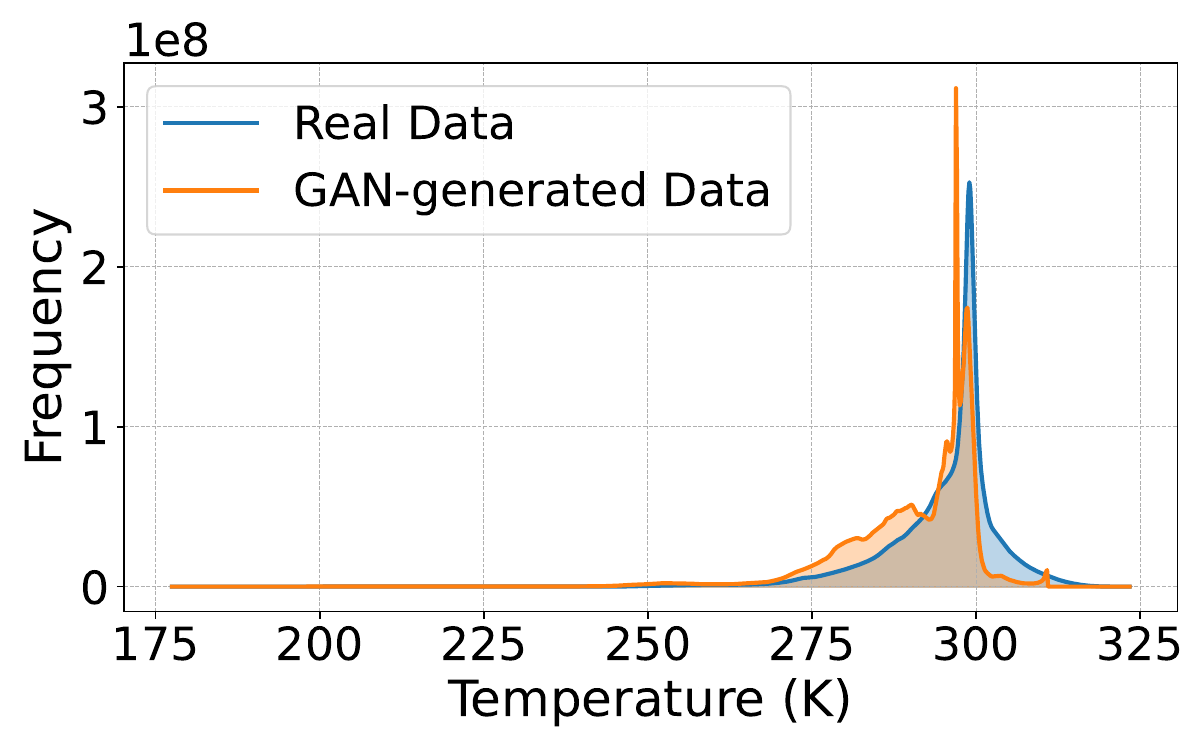}
    \caption{10K timestamps}
    \label{fig:distribution_10K}
  \end{subfigure}
  \hfill
  \begin{subfigure}[b]{0.33\textwidth}
    \centering
    \includegraphics[width=\linewidth]{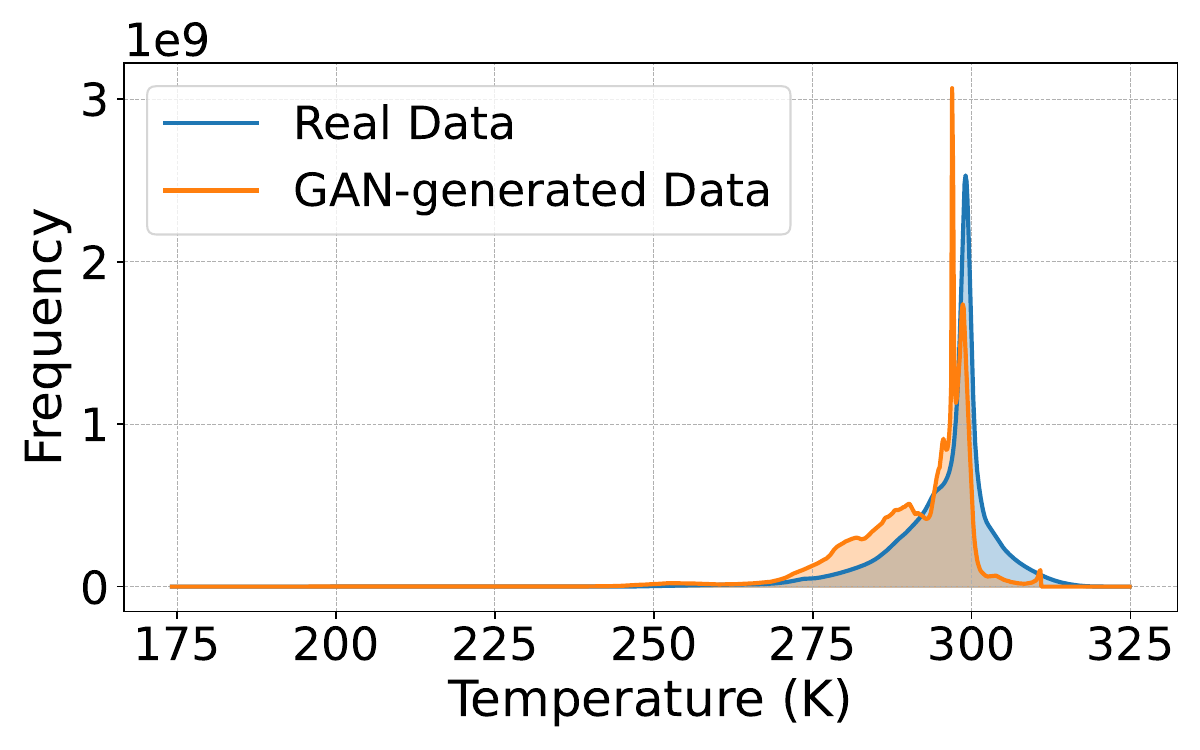}
    \caption{96K timestamps}
    \label{fig:distribution_96K}
  \end{subfigure}
  \caption{Distribution comparison between real and WGAN synthetic temperature data.}
  \label{fig:distribution_comparison_real_and_generator}
\end{figure*}

\stab
\noindent{\bf Statistical properties comparison.}
Figure~\ref{fig:distribution_comparison_real_and_generator} further illustrates the distribution of values for the real and WGAN-generated data. We present the comparison for three versions of the datasets, containing 500, 10K, and 96K timestamps, respectively. Observe that the similarity of the distributions is high. 

\stab
\stab
\noindent{\bf Benchmark performance testing.}
As a final test, we demonstrate that the compression ratios of WGAN-generated data with 500, 4K, 10K, and 96K timestamps closely align with those of the real data.
Table~\ref{table:final-cr-real-gan-redseaT} shows the results for 
three relative error bounds $\epsilon = 10^{-2}$, $10^{-3}$ and $10^{-4}$. Each group of rows contains the results of a real dataset and its corresponding synthetic one. Consistently with our results so far, {\GraphCom} is always the best method, whereas ZFP is always the worst. Although the exact values of the compression ratios are not identical between the real and WGAN generated data, the relative differences among the methods are largely preserved.


\begin{table}[t]
\caption{Compression ratios $\rho$ for real and synthetic data across multiple error bounds. \textbf{Bold} values indicate best performance; \underline{underlined} values are runner-ups.}
\label{table:final-cr-real-gan-redseaT}
\resizebox{\linewidth}{!}{
\begin{tabular}{lcrrrrr}
\hline
\bf  Dataset & $\epsilon$ & \bf SPERR & \bf ZFP  & \bf SZ 3.1 & \bf HPEZ  & \GraphCom \\
\hline\hline
\RS500    & $10^{-2}$ & 89.4 & 8.58 & \underline{77.87}  & 76.82 &  {\bf 99.74}   \\
              & $10^{-3}$ & 14.49 & 4.79 & 13.10   & \underline{13.39} &  {\bf 14.95}   \\
              & $10^{-4}$\vspace{4pt} & 5.89  & 2.99 & \underline{5.90}    & 5.73  &  {\bf 6.39}    \\ 
\WGAN500 & $10^{-2}$ & 34.81 & 5.37 & 39.09  & \underline{45.47} &  {\bf 64.25}   \\
              & $10^{-3}$ & 9.22  & 3.57 & \underline{11.05}  & 9.93  &  {\bf 13.05}   \\
              & $10^{-4}$ & 4.77  & 2.67 & \underline{5.48}   & 4.89  &  {\bf 5.98}    \\ \hline
\RS4K     & $10^{-2}$ & 89.75  & 8.44 & \underline{93.76}  & 92.17 &  {\bf 122.78}  \\
              & $10^{-3}$ & 13.97 & 4.13 & 14.06  & \underline{14.55} &  {\bf 17.18}   \\
              & $10^{-4}$\vspace{4pt} & 5.78  & 2.97 & \underline{5.97}   & 5.86  &  {\bf 7.14}    \\ 
\WGAN-4K  & $10^{-2}$ & 34.84 & 5.37 & 39.78  & \underline{45.61} &  {\bf 66.18}   \\
              & $10^{-3}$ & 9.22  & 3.57 & \underline{11.08}  & 9.89  &  {\bf 13.63}   \\
              & $10^{-4}$ & 4.76  & 2.68 & \underline{5.50}    & 4.93  &  {\bf 6.09}    \\ \hline
\RS10K    & $10^{-2}$ & 92.05 & 8.56 & \underline{98.66}  & 96.04 &  {\bf 128.83}  \\
              & $10^{-3}$ & 14.28 & 4.16 & 14.48  & \underline{15.07} &  {\bf 17.26}   \\
              & $10^{-4}$\vspace{4pt} & 5.86  & 2.99 & \underline{6.12}   & 5.87  &  {\bf 7.29}    \\
WGAN-10K & $10^{-2}$ & 28.44 & 5.97 & 53.21  & \underline{56.39} &  {\bf 83.61}   \\
              & $10^{-3}$ & 8.03  & 3.83 & 10.35  & \underline{11.71} &  {\bf 15.97}   \\
              & $10^{-4}$ & 4.37  & 2.59 & \underline{5.35}   & 5.28  &  {\bf 6.12}    \\ \hline
\RS96K    & $10^{-2}$ & 92.54 & 8.53 & \underline{105.21} & 102.59&  {\bf 141.75}  \\
              & $10^{-3}$ & 14.20  & 4.15 & 14.94  & \underline{15.29} &  {\bf 19.14}   \\
              & $10^{-4}$\vspace{4pt} & 5.85  & 2.99 & \underline{6.13}   & 5.95  &  {\bf 7.65}    \\ 
\WGAN96K & $10^{-2}$ & 40.40  & 5.97 & 54.78  & \underline{59.81} &  {\bf 90.86}   \\
              & $10^{-3}$ & 9.53  & 3.83 & 10.44  & \underline{11.86} &  {\bf 15.28}   \\
              & $10^{-4}$ & 4.78  & 2.59 & \underline{5.38}   & 5.31  &  {\bf 6.19}    \\ 
\hline
\end{tabular}}
\end{table}

\section{Conclusion}\label{sec:conclusion}
In this paper, we propose a novel graph-based error-bounded lossy compression method for scientific data, called \GraphCom.
At each timestamp, we convert the original grid data into graphs using several segmentation schemes inspired by image processing.
Then, we propose {\GNN}, a temporal graph autoencoder, to learn the latent representations of the graph data, effectively reducing the size of the original data.
The decoder of {\GNN} along with the segmentation are utilized to transform the latent representation back into grid data. The maximum point-wise relative error of the decompressed data is guaranteed to be bounded by a user-defined error bound.
Due to the confidentiality of our RedSea data, we train WGAN a adversarial generative model to generate and open-source synthetic versions of the dataset. 
Our experiments show that {\GraphCom} achieves superior compression ratios: for a relative error bound of $10^{-2}$, it compresses the temperatures in the RedSea data from 373.1GB down to 2.78GB. Moreover {\GraphCom} consistently outperforms the SOTA methods, SPERR, ZFP, SZ3.1, and HPEZ, for most tested cases, by a margin of 22\% up to 50\%.


\bibliographystyle{IEEEtran}
\bibliography{main}

\begin{IEEEbiography}[{\includegraphics[width=1.1in,height=1.3in,clip,keepaspectratio]{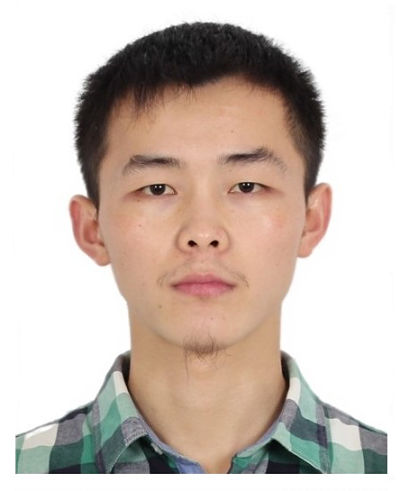}}]{Guozhong~Li}
is currently a postdoctoral research fellow at King Abdullah University of Science and Technology (KAUST). He obtained his Ph.D. degree from Hong Kong Baptist University (HKBU), and his MS and BS degree from the University of Electronic Science and Technology of China (UESTC), Chengdu, China. His research interests include scientific data compression, time series data mining, representation learning, and graph management.
\end{IEEEbiography}

\begin{IEEEbiography}[{\includegraphics[width=1.1in,height=1.3in,clip,keepaspectratio]{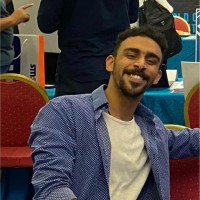}}]{Muhannad~Alhumaidi}
is currently a PhD student in King Abdullah University of Science and Technology (KAUST). His research interests include scientific data compression and large language model for data compression.
He obtained his MS from KAUST and his BS from Purdue University, both in Computer Science.
\end{IEEEbiography}

\begin{IEEEbiography}[{\includegraphics[width=1.1in,height=1.3in,clip,keepaspectratio]{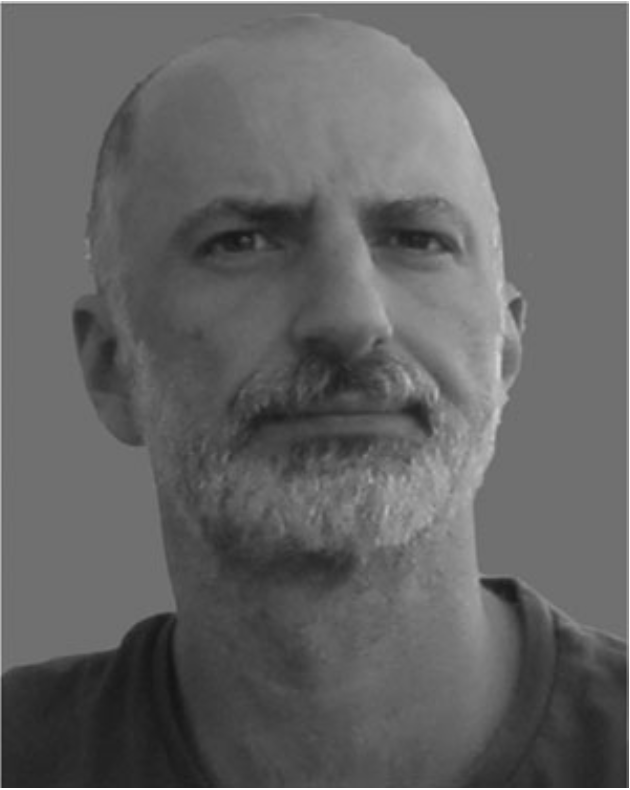}}]{Spiros~Skiadopoulos}
is a Professor with the Department of Informatics and Telecommunications, University of the Peloponnese and the director of the MSc program in Data Science. 
He received the PhD degree from the National Technical University of Athens (NTUA) and the MPhil degree from the Manchester Institute of Science and Technology (UMIST). He has served in the program committee of several venues and participated in various research and development projects. His scientific contribution received a large number of citations. For more information, please visit \url{https://users.uop.gr/~spiros/}.
\end{IEEEbiography}

\begin{IEEEbiography}[{\includegraphics[width=1.1in,height=1.3in,clip,keepaspectratio]{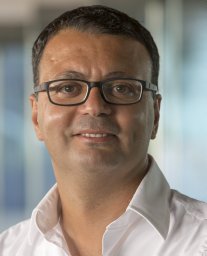}}]{Ibrahim~Hoteit}
is a Professor at KAUST. He leads the Climate Change Center, a national initiative supported by the Saudi Ministry of Environment, and directs the Aramco Marine Environment Center at KAUST.
His research focuses on studying the circulation and the climate of the Saudi marginal seas: the Red Sea and the Arabian Gulf, and understanding their impact on the ecosystems health and productivity. 
It involves effective use and integration of general circulation models with in-situ and satellite observations, including the development and implementation of data inversion, assimilation, and uncertainty quantification techniques suitable for large scale applications. 
\end{IEEEbiography}

\begin{IEEEbiography}[{\includegraphics[width=1.1in,height=1.3in,clip,keepaspectratio]{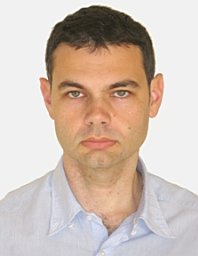}}]{Panos~Kalnis} is a Professor at KAUST and served as Chair of the Computer Science program. Before that, he was assistant professor at the National University of Singapore (NUS) and visiting assistant professor at Stanford University. He has served as associate editor for TKDE from 2013 to 2015, and on the editorial board of the VLDB Journal from 2013 to 2017. He received his PhD from the Hong Kong University of Science and Technology (HKUST) in 2002. His research interests include Big Data, Parallel and Distributed Systems, Large Graphs and Systems for Machine Learning. For more information, please visit \url{https://scholar.google.com/citations?user=-NdSrrYAAAAJ}
\end{IEEEbiography}

\end{document}